%% file: paper.tex
\documentclass[10pt,twocolumn,letterpaper]{IEEEtran}

\usepackage{times}
\usepackage{epsfig}
\usepackage{graphicx}
\usepackage{amsmath}
\usepackage{amssymb}
\usepackage{xspace}
\usepackage{rotating}
\usepackage{units}
\usepackage{multirow}
\usepackage{algorithm}
\usepackage{subfigure}
\usepackage{tikz}
\usepackage{dsfont}
\usepackage{booktabs}
\usepackage{lettrine}
\usepackage{array}
\usepackage{color}
\usepackage{grffile}
\usepackage{algorithmic}

\def \ie {\emph{i.e. }}

\input{macros.tex}

\begin{document}

\title{Interferences in match kernels}

\author{
    \IEEEauthorblockN{Naila Murray\IEEEauthorrefmark{1}, Herv\'e J\'egou\IEEEauthorrefmark{2}, Florent Perronnin\IEEEauthorrefmark{1}, Andrew Zisserman\IEEEauthorrefmark{3}}
    \\\IEEEauthorblockA{\IEEEauthorrefmark{1} \footnotesize{Xerox Research Centre Europe}
    \texttt{[firstname.lastname@xrce.xerox.com]}}
    \\\IEEEauthorblockA{\IEEEauthorrefmark{2}Facebook AI Research
    \texttt{[rvj@fb.com]}}
    \\\IEEEauthorblockA{\IEEEauthorrefmark{3}Department of Engineering Science, University of Oxford
    \texttt{[az@robots.ox.ac.uk]}}
}

\maketitle
\thispagestyle{empty}

\input{abstract.tex}


\input{introduction.tex}
\input{related.tex}
\input{matchkernels.tex}
\input{democratic.tex}

\input{gmp.tex}

\input{exp_retrieval.tex}

\section{Conclusion}

The key motivation for this paper is to reduce the interference between
local descriptors when aggregating them to produce a vector
representation of an image.
It is addressed by two novel and related aggregation techniques which aim to equalise the contribution of each descriptor to the final image-level representation.
The first aims to ``democratise" the contribution of each descriptor to a set comparison metric, while the second aims to equalise the similarity between each descriptor and the image-level representation.
This resulting representation compares favourably with the state-of-the-art in short- and mid-sized descriptors used in large-scale image search.
Future work will focus on the application of the proposed
aggregation mechanisms in deeper matching pipelines.

%-----------------------------------------------------------------------
\section*{Acknowledgments}
This work was done within the Project Fire-ID, supported by the ANR French research agency,
and also supported by ERC grant VisRec no.\ 228180.

\bibliographystyle{ieeetr}
\bibliography{egbib}

\begin{IEEEbiography}[{\includegraphics[width=1in,height=1.2in,clip,keepaspectratio]{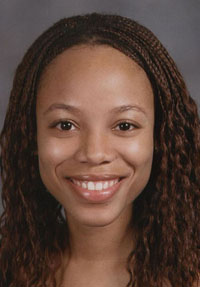}}]{Naila Murray}
obtained a BSc in electrical engineering from Princeton University in 2007, and MSc and PhD degrees from the Universitat Aut\`onoma de Barcelona in 2009 and 2012 respectively.
She joined Xerox Research Centre Europe in 2013, where she is currently a senior scientist and manager of the computer vision team. Her research interests include scene understanding, visual attention and image aesthetics analysis.
\end{IEEEbiography}

\begin{IEEEbiography}[{\includegraphics[width=1in,height=1.2in,clip,keepaspectratio]{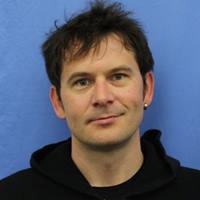}}]{Herv\'e J\'egou} is a former student of the Ecole Normale Superieure de Cachan, holding a PhD (2005) from University of Rennes defended on the subject of error-resilient compression and joint source channel coding. He  joined INRIA in 2006 as a researcher, where he turned to Computer Vision and Pattern Recognition. He recently joined Facebook AI Research as a research scientist.
\end{IEEEbiography}

\begin{IEEEbiography}[{\includegraphics[width=1in,height=1.2in,clip,keepaspectratio]{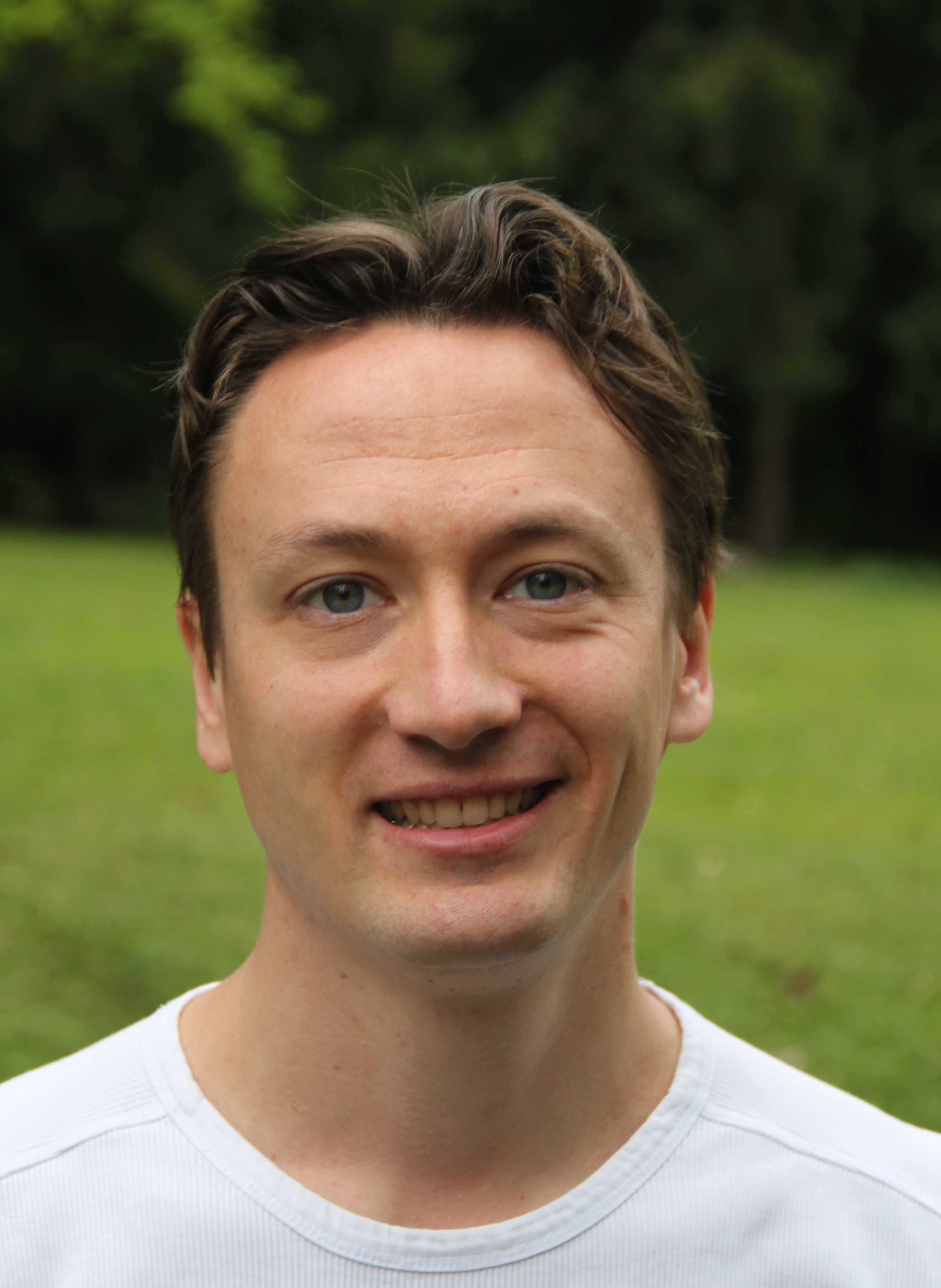}}]{Florent Perronnin} holds an Engineering degree from the Ecole Nationale Sup\'erieure des T\'el\'ecommunications de Paris and a Ph.D. degree from the Ecole Polytechnique F\'ed\'erale de Lausanne.
From 2000 to 2001, he was a speech research engineer at Panasonic in Santa Barbara, California.
From 2005 to 2015, he was with the Xerox Research Centre Europe (XRCE) in Grenoble, France, first as a research scientist, then as a principal scientist and manager of the computer vision team.
From May 2015 to June 2016 he was the manager of the Facebook AI Research lab in Paris.
Since July 2016, he is back at XRCE where he is the scientific lead of the analytics lab.
\end{IEEEbiography}

\begin{IEEEbiography}[]{Andrew Zisserman} is a professor of computer vision engineering in
the Department of Engineering Science at the University of Oxford.
\end{IEEEbiography}

\end{document}

%% file: macros.tex
\newcommand{\mypar}[1]{{\medskip \noindent \bf #1}}

\definecolor{ForestGreen}{RGB}{40,166,40}
\newcommand*{\DashedArrow}[1][]{\mathbin{\tikz [baseline=-0.25ex,-latex, dashed,line width=1pt,#1] \draw [#1] (0pt,0.5ex) -- (1.0em,0.5ex);}}
\newcommand*{\Arrow}[1][]{\mathbin{\tikz [baseline=-0.25ex,-latex, solid,line width=1pt,#1] \draw [#1] (0pt,0.5ex) -- (1.0em,0.5ex);}}
\newcommand*{\ThickArrow}[1][]{\mathbin{\tikz [baseline=-0.25ex,-latex, solid,line width=2.2pt,#1] \draw [#1] (0pt,0.5ex) -- (1.0em,0.5ex);}}

\newcommand{\card}[1]{\mathrm{card}({#1})}

\definecolor{darkgreen}{rgb}{0.0, 0.5, 0.0}

\newcommand\cmtg[1]{\textcolor{darkgreen}{#1}\xspace}

\def\eg{\emph{e.g.}\xspace}
\def\etal{\emph{et al.}\xspace}

\def\Cset{\mathcal C}
\def\Xset{\mathcal X}
\def\Yset{\mathcal Y}
\def\Kset{\mathcal K}
\def\Ksetn{\mathcal K^\star}

\def\bsigma{\boldsymbol{\sigma}}

\def\k{\mathsf{|\Cset|}}
\def\K{\mathbf{K}}

\def\D{D}

\def \x {\mathbf{x}}
\def \y {\mathbf{y}}

\def\1{\mathds{1}}

\def\vone{{\boldsymbol{1}}}

\newcommand\bphi{\boldsymbol{\phi}}
\newcommand\bPhi{\boldsymbol{\Phi}}
\def\triemb{\bphi_\triangle}
\def\fvemb{\bphi_{fv}}

\def\bpsi{\boldsymbol{\psi}}

\def\bxi{\boldsymbol{\xi}}

\def\psis{\bpsi_\mathrm{s}}
\def\psid{\bpsi_\mathrm{d}}
\def\psig{\bpsi_{\mathrm{gmp}}}
\def\gmp{\mathrm{gmp}}

\newcommand\balpha{\boldsymbol{\alpha}}
\newcommand\bA{\boldsymbol{A}}
\newcommand\bI{\boldsymbol{I}}
\newcommand\bM{\boldsymbol{M}}
\newcommand\bN{\boldsymbol{N}}
\newcommand\bQ{\boldsymbol{Q}}

\def\Re{\mathbb{R}}

\def\embname{T-embedding\xspace}

%% file: abstract.tex
%%%%%%%%% ABSTRACT
\begin{abstract}
We consider the design of an image representation that embeds and aggregates a set of local descriptors into a single vector. 
Popular representations of this kind include the bag-of-visual-words, the Fisher vector and the VLAD.
When two such image representations are compared with the dot-product, 
the image-to-image similarity can be interpreted as a match kernel.
In match kernels, one has to deal with {\em interference},
\ie with the fact that even if two descriptors are unrelated, 
their matching score may contribute to the overall similarity.

We formalise this problem and propose two related solutions,
both aimed at equalising the individual contributions 
of the local descriptors in the final representation. 
These methods modify the aggregation stage by including a set of per-descriptor weights.
They differ by the objective function that is optimised to compute those weights.  
The first is a ``democratisation'' strategy that aims at equalising the relative importance of each descriptor in the set comparison metric.
The second one 
involves
equalising the match of a single descriptor to the aggregated vector. 

These concurrent methods give a substantial performance boost over the state of the art in image search with short or mid-size vectors, 
as demonstrated by our experiments on standard public image retrieval benchmarks.
\end{abstract}

%-------------------------------

%% file: introduction.tex
% !TEX root = paper.tex
\section{Introduction}
\label{sec:introduction}

\begin{figure*}[!ht]
\centering
\subfigure[Sum aggregation]{
    \includegraphics[width=0.3\linewidth]{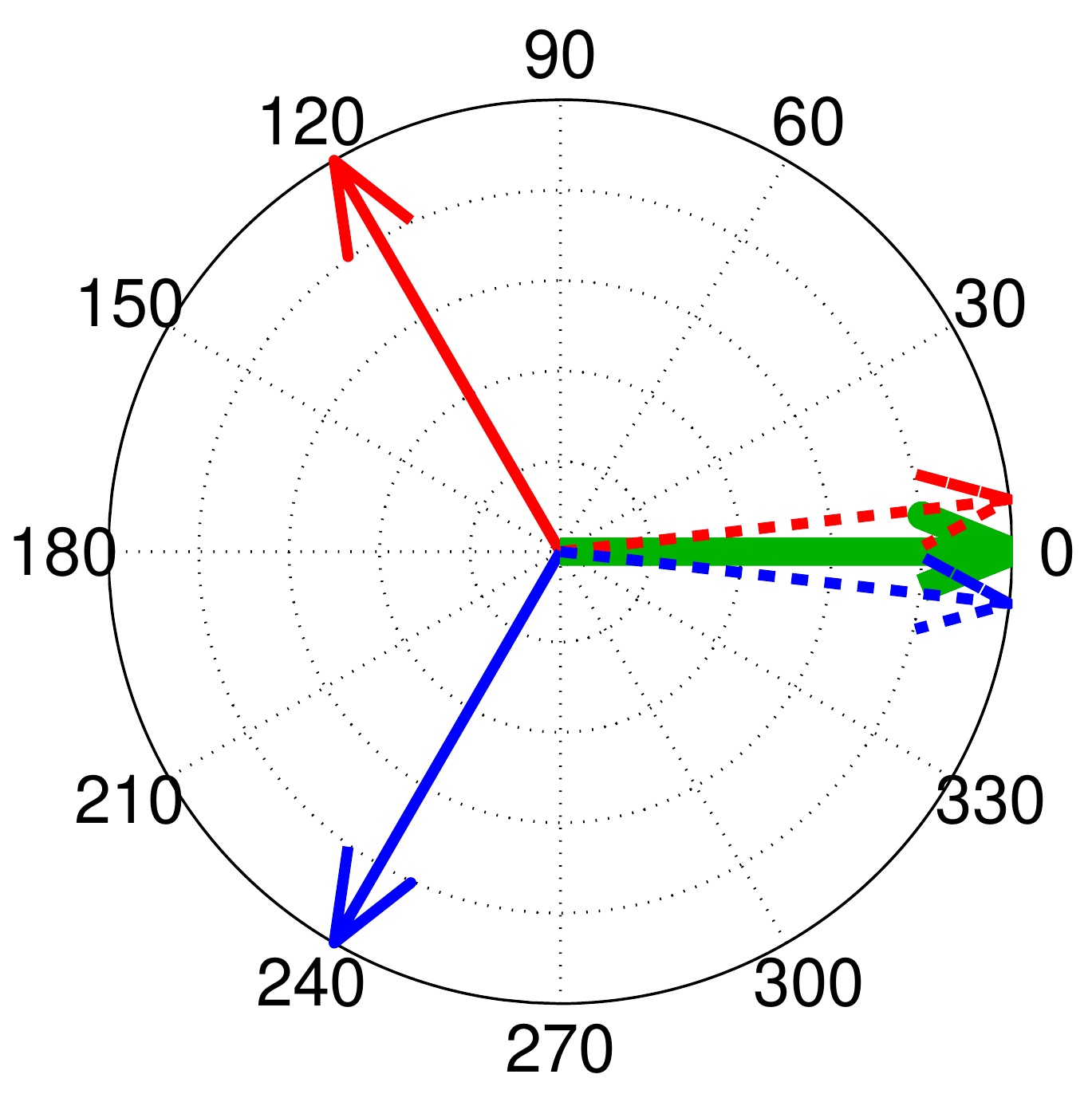}
    \label{fig:wgts_eg_sum}
}
\subfigure[DMK]{
    \includegraphics[width=0.3\linewidth]{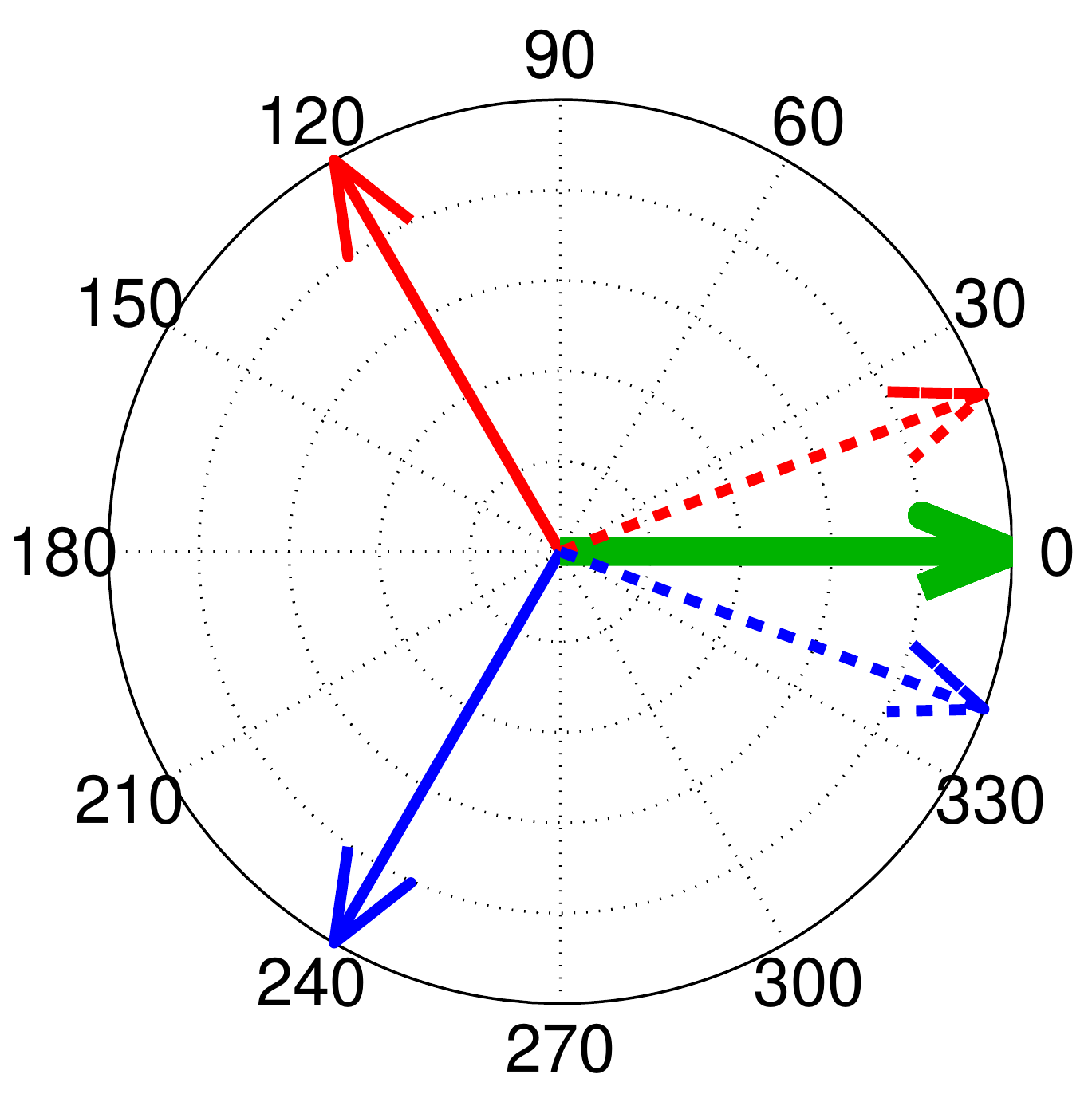}
    \label{fig:wgts_eg_dk}
}
\subfigure[GMP]{
    \includegraphics[width=0.3\linewidth]{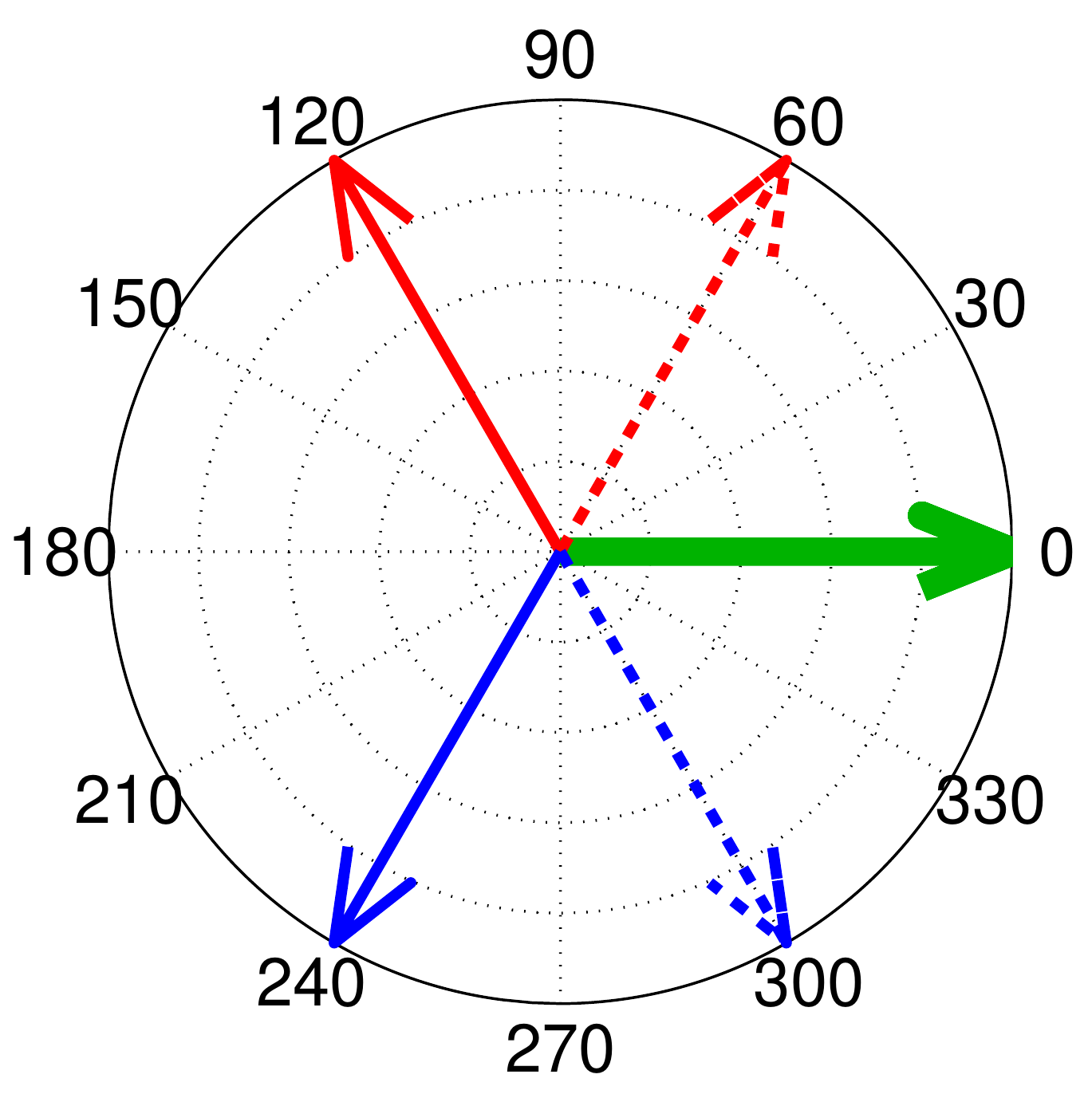}
    \label{fig:wgts_eg_gmp}
}
\caption{We show the effect of pooling a single descriptor encoding (\textcolor{red}{$\Arrow[->]$} or \textcolor{blue}{$\Arrow[->]$})
with a set of tightly-clustered descriptor encodings (\textcolor{ForestGreen}{$\ThickArrow[->]$}).
Two aggregated representations are shown: \textcolor{red}{$\Arrow[->]$} + \textcolor{ForestGreen}{$\ThickArrow[->]$} = \textcolor{red}{$\DashedArrow[->,densely dashed]$} 
and \textcolor{blue}{$\Arrow[->]$} + \textcolor{ForestGreen}{$\ThickArrow[->]$} = \textcolor{blue}{$\DashedArrow[->,densely dashed]$}. 
With sum aggregation (a), the cluster of descriptors \textcolor{ForestGreen}{$\ThickArrow[->]$} 
dominates the final representations \textcolor{red}{$\DashedArrow[->,densely dashed]$} 
and \textcolor{blue}{$\DashedArrow[->,densely dashed]$}, 
and as a result they are very similar to each other. 
With the proposed democratic (DMK) (b) and Generalised Max Pooling (GMP) (c) aggregations, 
both descriptors contribute meaningfully, resulting in more distinguishable pooled representations.
In this figure, we show the result of the non-regularised GMP.
However, in practice, we always use the regularised version in our experiments (with a fixed regularisation partameter $\lambda=1$)}
\label{fig:wgts_eg}
\end{figure*}

\lettrine{C}{onsider} the problem of large-scale instance-level image retrieval:
given an image of an object (\eg a Coca Cola can) or a landmark (\eg the Eiffel Tower), 
we are interested in retrieving images of the exact same object or landmark
in a potentially very large database of images.
The state-of-the-art for this task involves describing each image by a set of vectors 
(a bag-of-features), each feature representing a sub-part (a patch) of the considered image.
The similarity between two images is then measured by matching the similarity
between the descriptor sets~\cite{Low04,SiZ03}.
In this work, we focus on the scalable approach which involves aggregating patch-level
descriptors into a single fixed-length image-level descriptor.
The similarity between two images is measured as the similarity between
the two compounded descriptors using simple measures such as the dot-product, the Euclidean distance or the cosine similarity 
(which are all equivalent when dealing with $\ell_2$-normalised image descriptors as is standard practice).
This has been the dominating paradigm for very large-scale image retrieval because,
in combination with indexing techniques such as Product Quantisation (PQ)~\cite{JDS11},
it enables scaling to hundreds of millions of images on a single machine~\cite{JPDSPS12}.

Given a set of local descriptors, the standard pipeline to compute an image-level representation involves two main steps:
the {\em embedding step}, in which an embedding function $\bphi$ individually maps each vector of the set into a high-dimensional space;
and an {\em aggregation step}, in which an aggregation function $\bpsi$ produces a single vector from the set of mapped vectors, 
for instance using sum- or max-pooling\footnote{
The embedding step is closely related to the \emph{coding} step as usually considered in the literature, 
and the aggregation step is close to the \emph{pooling} step. 
We use another terminology to avoid confusion, because in our case all the operations
applied on a per descriptor basis are included in the embedding step.
For instance, geometry-based pooling such as spatial pyramid pooling~\cite{LSP06}
is included in the embedding step in our formulation.
Consequently, the output dimensionality of $\bphi$ is typically the same
as that of the final vector representation of the set.}.

We note that the first step -- the embedding step -- has received much attention in the last decade.
This has consequently led to a flurry of image-level representations including
the bag-of-visual-words (BOW)~\cite{SiZ03}, 
BOW with multiple-assignment~\cite{JDS10a,JSHV10} or soft-assignment~\cite{PCISZ08,GVSG10},
locality-constrained linear coding~\cite{WYY10}, the Fisher Vector (FV) \cite{PD07,PSM10} or
the VLAD~\cite{JPDSPS12} -- see Section \ref{sec:rel:phi}.

Comparatively, much less work has been devoted to the second step -- the aggregation step.
The most popular aggregation mechanism involves summing the descriptor embeddings \cite{SiZ03,CDFWB04,PD07,BS09,JPDSPS12,ZYZH10}.
In such a case, the compounded representation of a set $\Xset$ simply writes as 
\begin{equation}\sum_{\x \in \Xset} \bphi(\x). \end{equation}
This summing strategy is popular because it is simple to implement and because it can be used
in conjunction with any embedding function.

However, a problem that arises when summing all the descriptor embeddings 
is that unrelated descriptors produce {\em interference}.
Indeed, if one adopts the dot-product similarity,
then given two sets $\Xset$ and $\Yset$,
their similarity $\Kset(\Xset,\Yset)$ can be rewritten as 
\begin{equation}
\label{eqn:smk}
\Kset(\Xset,\Yset) = \sum_{\x \in \Xset} \sum_{\y \in \Yset} \bphi(\x)^\top \bphi(\y),
\end{equation}
\ie the sum of pairwise similarities between {\em all} descriptors.
Consequently, unrelated patches may contribute to the kernel even if they have a low similarity. 
This is in contrast to approaches which perform patch-level matching~\cite{Low04}, 
in which only the strong matches (inliers) are counted to produce a similarity score.  
The interference issue is exacerbated when an image contains bursty descriptors~\cite{JDS09a},
\ie descriptors which though not necessarily identical, 
together form a mode in descriptor space (\eg window patches in an image of a building's facade as shown on Figure~\ref{fig:map_egs}).
Indeed, the cross-terms between such descriptors tend to dominate the similarity in equation \eqref{eqn:smk}. 
However, such bursty descriptors are not necessarily the most informative ones.

The interference issue can be addressed -- at least partly -- through the choice 
of a suitable embedding function $\bphi$ by ensuring that the similarity $\bphi(\x)^\top \bphi(\y)$
drops rapidly as $x$ and $y$ get dissimilar -- see Section \ref{sec:rel:phi}.
In this work, we take the embedding function $\bphi$ for granted and focus on the aggregation function $\bpsi$.
Our goal is to propose a generic framework which is applicable to any embedding function $\bphi$.
We do so by assigning one weight to each local descriptor to ``equalise'' the matching contribution of each descriptor.
We describe two instantiations of this generic principle.
First, we propose an aggregation strategy that tends to 
``democratise'' or equalise the contributions of each vector to the final similarity score between two sets.
This involves an optimisation problem which is solved  with a modified Sinkhorn algorithm~\cite{K08,S64}. 
The second criterion is a generalisation of the max-pooling aggregation strategy.
It enforces the similarity between a single patch descriptor
and the aggregated representation to be constant.
This optimisation problem can be cast as a ridge regression problem
for which a simple closed form formula exists.

Figure~\ref{fig:wgts_eg} shows results with sum aggregation, democratic aggregation and generalised max pooling aggregation for an example involving  2D descriptors.
Aggregation with $\psis$ results in representations which are highly influenced by the bursty descriptors (shown as green vectors) and not discriminative.
For both $\psid$ and $\psig$, the aggregated representations are more influenced by the rare 
but potentially more informative descriptor and are thus more distinguishable with respect to one another.

The remainder of this paper is organised as follows.
Section~\ref{sec:litreview} reviews related works while Section~\ref{sec:background} 
introduces notations and motivates our contributions.
Section~\ref{sec:democratic} and Section~\ref{sec:gmp} introduce our methods. 
The experiments are presented in Section~\ref{sec:experiments}. 
They show that our methods have a significant improvement over previous techniques, 
and are also complementary to the so-called power-law normalisation~\cite{PSM10}.

This paper extends the publications~\cite{JZ14,MP14} in the following ways:
it contains expanded related work and motivation sections;
it presents in a unified way the democratic and generalised max pooling algorithms
thus highlighting their differences and commonalities; and
it provides additional results on the task of instance-level image retrieval.

%% file: related.tex
% !TEX root = paper.tex
\section{Related Work}
\label{sec:litreview}

We now review those works which deal with the interference problem in match kernels
of the form provided in equation \eqref{eqn:smk}.
Note that alternative similarities such as the sum-max ``kernel'' proposed by Walvaren \etal~\cite{WCG03}
are not considered in this section because, as shown by Lyu~\cite{Lyu05},
they cannot be rewritten as the dot-product between two compounded vectors 
(\ie the defined similarity is not a Mercer kernel).
Consequently, such similarities are not straightforwardly applicable in the very large-scale setting.

As mentioned earlier, it is possible to deal with interferences in match kernels either at the embedding stage,
\ie by choosing an appropriate function $\bphi$,
or at the aggregation stage, \ie by choosing an appropriate function $\bpsi$.
The two following sections review these two alternatives respectively.

\subsection{Dealing with interferences at the embedding stage}
\label{sec:rel:phi}

The interference problem can be addressed, at least in part, at the embedding stage.
Indeed, by embedding the low-level descriptors into a higher-dimensional space,
most embedding functions ensure that, for any pair of vectors $(\x,\y)$ describing two patches, 
the similarity $\bphi(\x)^\top \bphi(\y)$ is large if the patches match, 
and close to zero if they do not, \ie the magnitude of $\bphi(\x)^\top \bphi(\y)$ is small for unrelated patches. 

For instance, in the BOW framework~\cite{SiZ03,CDFWB04},
descriptors contribute by a constant to the similarity if they are assigned to the same cell, and by 0 otherwise.
An issue however with such an approach is that of quantisation: 
two very similar descriptors may be assigned to two different cells and therefore may not contribute to the final similarity.
Hence, several improvements to the BOW have been proposed.
This includes the assignment of a descriptor to multiple cells~\cite{JHS07,PCISZ08},
the soft-assignment of descriptors to cells~\cite{WCM05,PDC06,GVSG10},
the use of multiple vocabularies~\cite{JC12} or sparse coding~\cite{MBP08,YYGH09,BBLP10,WYY10}.
All the previous methods are less prone to quantisation noise while keeping
the BOW property that two dissimilar patches should have a zero (or very close to zero) similarity.

To reduce interferences in the BOW framework,
the standard approach is to increase the size of the visual codebook.
Indeed, as the dimensionality of the codebook increases,
the probability of collision between two different patches,
\ie the probability that they are assigned to the same quantisation cell, decreases.
However, the probability that similar patches end-up in different cells also increases.
Hence, as the codebook size increases, two similar images are more likely to be deemed dissimilar.
To increase the embedding dimensionality without increasing the codebook size,
an alternative involves encoding higher-order statistics.
Such techniques include the Fisher Vector (FV)~\cite{PD07,PSM10,SPMV13}
which encodes the gradient of the log-likelihood of the data with respect to the parameters of a generative model, 
the Vector of Locally Aggregated Descriptors (VLAD)~\cite{JDSP10} which can be interpreted as a non-probabilistic
version of the FV~\cite{JPDSPS12} or the Super Vector (SV)~\cite{ZYZH10} which can be interpreted as the concatenation of a BOW and a VLAD.
The triangulation embedding~\cite{JZ14}, although closely related to the VLAD, introduces normalisation and 
dimensionality reduction steps which significantly improve its matching performance.

Codebook-free techniques have also been proposed.
This includes the Efficient Match Kernel (EMK) 
which involves the projection of local descriptors onto random Gaussian directions
followed by a cosine non-linearity~\cite{BS09},
or second-order pooling which involves encoding second order statistics of the patches~\cite{CCB12}.
Note that the latter technique can be interpreted as the raising of the patch-to-patch similarity to a power
factor which explicitly reduces the interference between dissimilar patches~\cite{Lyu05}.

Finally, geometrical encoding techniques such as Spatial Pyramid pooling~\cite{LSP06}
tend to reduce interference by reducing the size of the pooling region, which can be viewed
equivalently as embedding the data in a $P$ times higher dimensional space where $P$ denotes
the number of pooling regions.

\subsection{Dealing with interferences at the aggregation stage}

While the choice of a proper embedding function $\bphi$ may reduce the influence of interferences
in match kernels, it cannot address the problem of duplicate (or near-duplicate) patches which
have identical (or near-identical) embeddings.
In such a case, % the only way -> rvj: very dangerous to say this is the only way
we consider a way to reduce the undesirable influence of these 
bursty descriptors in the aggregation stage.
We note that in our formulation the function $\bpsi$ includes
both the compounding stage itself -- which we will refer to as pooling for simplicity in the next paragraph -- and any subsequent normalisation
on the image-level representations.
We therefore review related works on local descriptor pooling
and image-level descriptor normalisation.
\medskip

\paragraph{Descriptor pooling}
The term ``pooling" is used to refer to an operation which either
i) groups several local descriptors based on some similarity criterion or
ii) aggregates the descriptors into a single representation.
In this section, we use the second meaning.
On the one hand, pooling achieves some invariance to perturbations of the descriptors.
On the other hand, it leads to a loss of information.
Pooling is typically achieved by
either summing/averaging or by taking the maximum response. 

Sum-pooling has been used in many biologically-inspired visual recognition systems
to approximate the operation of receptive fields in early stages of the visual cortex
\cite{FM82,PCD08,JCD12}.
It is also a standard component in convolutional neural networks \cite{LBDHHHJ89}.
A major disadvantage of sum-pooling is that it is based on the incorrect assumption
that the descriptors in an image are independent
and that their contributions can be summed~\cite{BPL10,CVS12,SPMV13}.

Max-pooling was advocated by Riesenhuber and Poggio
as a more appropriate pooling mechanism for higher-level visual processing
such as object recognition \cite{RP99}.
It has subsequently been used in computer vision models of object recognition \cite{SWP05} and especially in neural networks \cite{RBL07,LGRN09}.
It has also recently found success in image classification tasks when used in conjunction with sparse coding techniques
\cite{YYGH09,BBLP10,GTCZ10,WYY10,BPL10}.
A major disadvantage of max-pooling is that it only makes sense when applied to embedding functions that encode a strength of association between a descriptor and a codeword, as is the case of the BOW
and its soft- and sparse-coding extensions.
However, it is not directly applicable to those representations which compute higher-order statistics
such as the FV or triangular embedding \cite{JZ14}.

Several extensions to the standard sum- and max-pooling frameworks have been proposed.
Koniusz \textit{et al.} introduced a pooling technique for BOV descriptors which sum-pooled only the largest activations per visual word based on an empirical threshold \cite{koniusz2013comparison}.
One can also transition smoothly from sum- to max-pooling using $\ell_p$- or softmax-pooling \cite{BPL10}, or use ``mix-order" max pooling to incorporate some frequency information into a max-pooling framework \cite{liu2011defense}.
It is also possible to add weights to obtain a weighted pooling.
J\'egou \textit{et al.} proposed several re-weighting strategies for addressing visual burstiness 
\cite{JDS09a}.
These include penalising multiple matches between a query descriptor and a database image
and penalising the matches of descriptors which are matched to multiple database images
(\ie IDF weighting applied at the descriptor level rather than the visual word level).
Torii \textit{et al.} proposed another re-weighting scheme for BOW-based representations,
which soft-assigns to fewer visual words those descriptors which are extracted from a repetitive
image structure \cite{TSPO13}.
De Campos \textit{et al.} computed weights using a saliency estimation method trained on external data in order to cancel-out the influence of irrelevant descriptors \cite{dCCP12}.
While our democratic or generalised max pooling can be viewed as an instance of weighted pooling (see section \ref{sec:wgh}),
two major differences with previous re-weighting schemes are that our weights are computed on a per-image basis, {\em i.e.} independently of other database images or other external information, and that the weights serve a different purpose: to adjust the influence of frequent and rare descriptors.
\medskip

\paragraph{Image-level descriptor normalisation}

Many works make use of sum-pooling and correct \textit{a posteriori} for the incorrect independence assumption through normalisation of the pooled representation.
Arandjelovi\'{c} \textit{et al.} showed that, for the VLAD representation \cite{JPDSPS12},
applying $\ell_2$-normalisation to the aggregated representation of each pooling region mitigates the burstiness effect \cite{AZ13}.
Delhumeau \etal found that, for VLAD, $\ell_2$-normalising the descriptor residuals and then applying PCA before pooling was beneficial
\cite{DGJP13}.
Power normalisation has also been shown to be an effective heuristic for treating
frequent descriptors in BOW, FV or VLAD representations \cite{boughorbel2005generalized,PSL10,PSM10,JPDSPS12}.
While the square-rooting (\ie the power-normalisation with power parameter 0.5)
of histogram representations can be interpreted as a variance stabilising transform~\cite{FN04,WCM05},
such a power-normalisation scheme is largely heuristic in the case of the FV or the VLAD.
In general, these \textit{a posteriori} normalisations are either restricted to image representations based on a finite vocabulary and hence not applicable to
codebook-free representations such as the EMK, or are heuristic in nature.

One of the rare works which considered the independence problem in a principled manner
is that of Cinbis \etal which proposes a latent model to take into account inter-descriptor dependencies \cite{CVS12}.
However, this work is specific to representations based on Gaussian Mixture Models.
In contrast, our weighted pooling framework is generic and applicable to any aggregation-based representation.

%% file: matchkernels.tex
% !TEX root = paper.tex

\section{Match kernels and interferences}
\label{sec:background}

This section introduces match kernels (from both a dual and a primal view), 
their notations, and the properties that motivate the methods presented in the subsequent sections. 

\subsection{The dual view}

Let us consider two sets $\Xset$ and $\Yset$ such that
$\card{\Xset}=n$ and $\card{\Yset}=m$. 
Each set consists of a set of vectors, such as local descriptors extracted from an image.
We denote $\Xset = \{\x_1, \ldots, \x_n\}$ and $\Yset = \{\y_1, \ldots, \y_m\}$.
The elements $\x_i$ and $\y_j$ of $\Xset$ and $\Yset$ take their values in $\Re^d$ (\eg $d=$128 in the case of SIFT descriptors).
We denote by $\Re^d_*$ the space of sets of $d$-dim vectors such that $\Xset,\Yset \in \Re^d_*$.
We first consider match kernels, in a framework derived from Bo and Sminchisescu~\cite{BS09}\footnote{A minor difference with this prior work is that it normalises the vector representation of the set by the number of features.}, that have the form
\begin{equation}
\Kset(\Xset,\Yset) = \sum_{\x \in \Xset} \sum_{\y \in \Yset} k(\x,\y), 
\label{equ:matchkernel}
\end{equation}
where $k(x,y)$ is a kernel between individual vectors of the sets.
The match kernel is also written in matrix form as
\begin{equation}
\label{eqn:mkdual}
\Kset(\Xset,\Yset) = \vone_n^\top \, \K(\Xset,\Yset) \, \vone_m, 
\end{equation}
where $\vone_n=\underbrace{[1,\dots,1]}_{\times n}$, and we define the 
$n \times m$ matrix 
\begin{equation}
\K(\Xset,\Yset) = \left[ 
\begin{array}{ccc}
k(x_1,y_1) & \hdots & k(x_1,y_m) \\
\vdots     & \ddots & \vdots  \\
k(x_n,y_1) & \hdots & k(x_n,y_m) 
\end{array}
\right].
\end{equation}
This matrix contains all the pairwise similarities between the local descriptors of two images. 
For any kernel $\Kset$, we denote its normalised counterpart
\begin{equation}
\Ksetn(\Xset,\Yset) = \nu(\Xset)\ \nu(\Yset)\ \Kset(\Xset,\Yset),
\label{eqn:matchkernelnorm}
\end{equation} 
where the normalizer $\nu(.)$ is defined such that $\Ksetn(\Xset,\Xset)=\mathsf{1}$, \ie, $
\nu(\Xset)=\Kset(\Xset,\Xset)^{-1/2}. 
$%\end{equation}

\subsection{The primal view}
If $k(.,.)$ is a true Mercer kernel, there exists an explicit feature map
such that $k(x,y) = \bphi(x)^\top \bphi(y)$.
We focus on finite-dimensional embeddings $\bphi: \Re^d \rightarrow \Re^D$.
Typically, the size $D$ of the output space is significantly higher than that of the input space $d$,
\eg $D$ is on the order of $10^4$ or $10^5$.
We can thus rewrite Equation \eqref{equ:matchkernel} as:
\begin{eqnarray}
\Kset(\Xset,\Yset) = & \sum_{\x \in \Xset} \sum_{\y \in \Yset} \bphi(\x)^\top \bphi(\y) \nonumber \\
                   = &  \left( \sum_{\x \in \Xset} \bphi(x) \right)^\top \left( \sum_{\y \in \Xset} \bphi(y) \right).
\end{eqnarray}
If we denote 
\begin{equation}
\label{eqn:emXs}
\bxi(\Xset) = \sum_{\x \in \Xset} \bphi(x)
\end{equation}
then $\bxi: \Re^d_* \rightarrow \Re^D$ is the explicit feature map of the kernel $\Kset$:
\begin{equation}
\Kset(\Xset,\Yset) = \bxi(\Xset)^\top \bxi(\Yset).
\end{equation}
We can thus divide the construction of the feature map $\bxi(\Xset)$ into two steps,
namely \emph{embedding} and \emph{aggregation}. 
The embedding step maps each $x \in \Xset$ as
\begin{align}
%  \nonumber \\
\x       & \mapsto \bphi(\x). 
\label{equ:embedding}
\end{align}
The aggregation step computes a single vector from the set 
of embedded vectors $\bphi(\Xset) = \{\bphi(\x_1),\dots,\bphi(\x_n)\}$ through a function $\bpsi: \Re^D_* \rightarrow \Re^D$:
\begin{equation}
\label{eqn:emX}
\bxi(\Xset) = \bpsi\left(\bphi(\Xset)\right).
\end{equation}
When the aggregation function is a simple summation as is the case in Equation \eqref{eqn:emXs},
the aggregation function is denoted $\psis$.
Let $\bPhi_\Xset$ be the $D \times n$ matrix of patch embeddings for set $\Xset$:
$\bPhi_\Xset = [\bphi(\x_1), \ldots, \bphi(\x_n)]$. In matrix form, we have
\begin{equation}
\bxi(\Xset) = \bPhi_\Xset \vone_n.
\end{equation}

We now highlight the limitations of $\psis$.

\subsection{Interferences in match kernels}

The explicit feature map $\bxi$ of Equation (\ref{eqn:emX}) has the advantage of producing a
vector representation which is compatible with linear algebra, efficient linear classification 
and quantisation, to mention but a few relevant procedures.
However, this feature map gives an undue importance to bursty descriptors. 
To see this more clearly, it is enlightening to compare $\xi(\Xset)$ to itself.
In such a case, the contribution of a given vector $x$ to the overall similarity 
$\bxi(\Xset)^\top \bxi(\Xset)$ is given by $\bphi(x)^\top \psis(\Xset)$:
\begin{equation}
  \bphi(\x)^\top \sum_{\x' \in \Xset}  \bphi(\x') 
  = \|\bphi(\x) \|^2 + \bphi(\x)^\top  \sum_{\x' \in \Xset \setminus \x}  \bphi(\x'). 
  \label{equ:artifacts}
\end{equation}

The left term $\|\bphi(\x) \|^2$ isolates the similarity of the descriptor to itself. 
When the embedded descriptors are $\ell_2$-normalised, whether exactly as is the case in~\cite{DGJP13,JZ14} or approximately as is the case in~\cite{BS09},
then this first term is equal to 1 (exactly or approximately).

The term on the right can be interpreted as the ``noise'' polluting the contribution of $\x$ due to its interaction with the other vectors.
We consider two cases.
In the first case the vector $\x$ is ``unique'' in the sense that it is dissimilar to all other vectors: $\bphi(\x)^\top \bphi(x') \ll 1$ for all $x' \in \Xset \setminus \x$.
However, because the set $n$ may be large, the quantity $\sum_{\x' \in \Xset \setminus \x}$ may end up being comparable to the first term.
In the second case, $\x$ is bursty in the sense that there exist several vectors $\x'\in \Xset \setminus \x$ such that $\bphi(\x)^\top \bphi(\x') \approx 1$.
In this case, the contribution of $\x$ to the final similarity will be higher than that of a ``unique'' descriptor although, 
as noted earlier, a bursty descriptor is not necessarily more informative.

\subsection{Proposed solution}
\label{sec:wgh}

Given a set $\Xset$, we propose to introduce for each $\x \in \Xset$ a weight denoted $\alpha_\Xset(\x)$
which depends only on the element $\x$ and the set $\Xset$.
The purpose of these weights is to equalise in some sense the contribution of each element in the set to the final similarity score.
In such a case, the aggregated representation writes as:
\begin{equation}
\bxi(\Xset) = \sum_{\x \in \Xset} \alpha_\Xset(\x) \bphi(\x),
\end{equation}
or in matrix form:
\begin{equation}
\bxi(\Xset) = \bPhi_\Xset \balpha_\Xset.
\end{equation}
where $\balpha_\Xset$ denotes the $n$-dimensional vector of weights for set $\Xset$.
In this case, the similarity between $\Xset$ and $\Yset$ rewrites as:
\begin{align}
\Kset(\Xset,\Yset)  & = \left( \sum_{\x \in \Xset} \alpha_{\Xset}(\x) \bphi(x) \right)^\top 
   \left(\sum_{\y \in \Yset} \alpha_{\Yset}(\y) \bphi(\y) \right) \\
& = \sum_{\x \in \Xset} \sum_{\y \in \Yset} \alpha_{\Xset}(\x) \, \alpha_{\Yset}(\y) \, \bphi(\x)^\top \bphi(\y) \\
& = \balpha_\Xset^\top \K(\Xset,\Yset) \balpha_\Yset.
\end{align}
Note that the previous equation is equivalent to defining a new match kernel
\begin{equation}
\kappa(\x,\y)=\alpha_{\Xset}(\x) \, \alpha_{\Yset}(\y) \, k(\x,\y).
\end{equation} 

We now propose in the two following sections
two different objective functions to compute the weights $\alpha$.

%% file: democratic.tex
% !TEX root = paper.tex

\section{Democratic aggregation}
\label{sec:democratic}

We define that a kernel $\Kset$ over sets is \emph{democratic} if and only if, 
for any set $\Xset$ s.t. $\card{\Xset}=n$, it satisfies
\begin{equation}
  \label{eqn:demprop}
\bphi(\x)^\top \sum_{\x' \in \Xset} \bphi(x') = C, \forall \x \in \Xset
\end{equation}
where the scaling factor $C$ may (or may not) depend on $\Xset$.
Note that the value of the constant $C$ has no influence given that the
final embedding $\bxi(\Xset)$ is typically $\ell_2$-normalised.
In other words, a democratic kernel ensures that all the vectors in $\Xset$ contribute equally to the self-similarity. 
Introducing the notation $\K(\Xset,\Xset)=\K_\Xset$, in matrix form equation \eqref{eqn:demprop} becomes
\begin{equation}
\K_\Xset \, \vone_n = C \, \vone_n.
\end{equation}

In the rest of this section, 
we present the optimisation problem aiming at producing a democratic kernel from an arbitrary one in the aggregation stage. 
Then we discuss the practical computation of the ``democratic'' weights and especially
a strategy to achieve convergence. 
Finally, we discuss the relationship between the democratic kernel and the Hellinger kernel in the case of the BOW embedding.
In what follows, whenever there is no ambiguity, we drop the subset $\Xset$ from the notations. 
In particular, $\K_\Xset$ simplifies to $\K$, $\bPhi_\Xset$ to $\bPhi$ and $\balpha_\Xset$ to $\balpha$.

\subsection{Democratic formulation} 
A kernel as defined in (\ref{equ:matchkernel}) is generally not democratic.
To turn a generic kernel into a democratic one,
we introduce a set of weights $\alpha$ to enforce the property
\begin{equation}
\alpha(\x) \bphi(\x)^\top \sum_{\x' \in \Xset}  \alpha(\x') \bphi(x') = C, \forall \x \in \Xset
\end{equation}
under the constraint $\forall \x \in \Xset, \alpha(x) > 0$. 
This rewrites in kernel form as
\begin{equation}
\alpha(\x) \sum_{\x' \in \Xset}  \alpha(\x') \, k(x,x') = C, \forall \x \in \Xset. 
\end{equation}
The problem is summarised as finding a matrix $\bA=\mathrm{diag}(\balpha)$ whose diagonal is strictly positive
and such that
\begin{equation}
\bA \K \bA \vone_n = C \vone_n. 
\label{eqn:dempb}
\end{equation}

\subsection{Computing democratic weights in practice}
\label{sec:demcomp}

We now describe how we approximately solve Equation \eqref{eqn:dempb} 
with a modified version of the Sinkhorn algorithm. Then we discuss the cost incurred
by this algorithm.

\mypar{Modified Sinkhorn scaling algorithm.}
It is worth noticing that equation \eqref{eqn:dempb} resembles that of projection to
a doubly stochastic matrix~\cite{S64}.
It is equivalent if $C=1$ and $\K$ is positive.  
Under additional assumptions (matrix $\K$ has total
support and is fully indecomposable~\cite{K08}), the Sinkhorn
algorithm converges to a unique solution satisfying
$\forall \x \in \Xset,\ \alpha(\x) > 0$.
It is a fixed-point algorithm that proceeds by alternately normalising the rows and columns.
We adopt a symmetric variant analysed by Knight~\cite{K08} and weaken the impact of each
iteration, as recently suggested~\cite{JSHV10}, by using a power
exponent smaller than $0.5$ for a smoother convergence.
The pseudo-code of this optimisation method is provided in Algorithm 1. 

\def\nbiter{$\mathrm{n}_\mathsf{iter}$}
\def\ones{$1$}
\begin{algorithm}[t]
{
\begin{tabular}{p{0.21\linewidth}p{0.88\linewidth}}

\textbf{Input:}  & Gram matrix $\K$      \hspace{0.7cm}      \cmtg{\ \ \ \% of size $n \times n$} \\
                 & parameters $\gamma$ and \nbiter \\
\textbf{Output:} & Weight vector $\balpha$ \\

\textbf{Initialisation:} & $\balpha = \vone_n$  
\end{tabular}
\smallskip 

\textbf{For} i=1 to \nbiter  \\
{\setlength{\parindent}{0.7cm} 
\noindent 
\indent $\bsigma = \mathrm{diag}(\balpha) \times \K \times \mathrm{diag}(\balpha) \times \vone_n$ 
\cmtg{\ \ \% Sums of rows}

\indent $\forall i,\ \alpha_i := \alpha_i / \sigma_i^\gamma $ 
\cmtg{\hspace{2.7cm} \% Update} \
}
}
\caption{Pseudo-code: computing democratic weights.
\label{alg:sinkhorn}}
\end{algorithm}

Sinkhorn is an algorithm that converges quickly. 
We stop it after 10 iterations for efficiency reasons.
Experimentally, no benefit comes from using more iterations.

\begin{figure}[t]
\includegraphics[width=\linewidth]{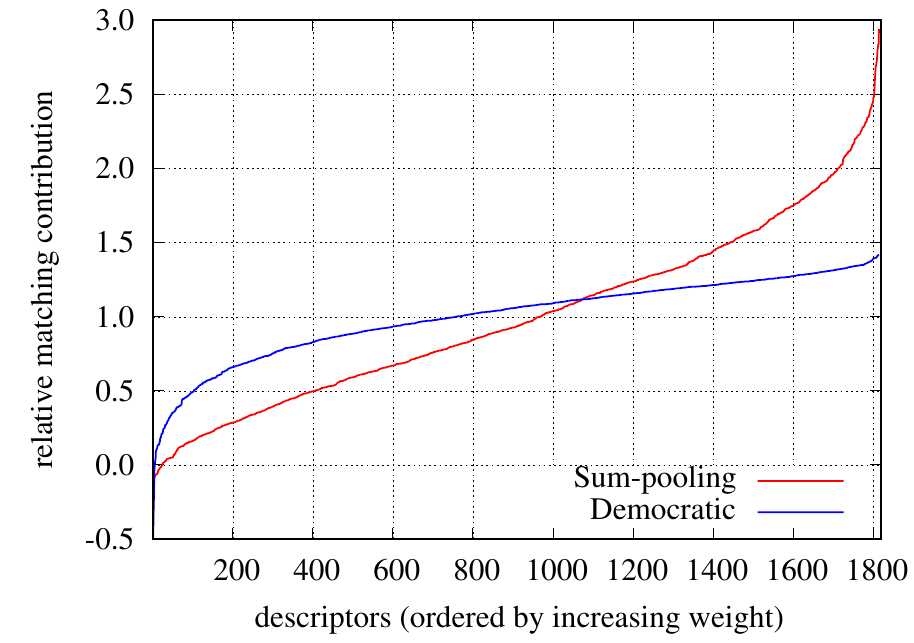}
\caption{Relative contribution of T-embedded descriptor in the match kernel (self-similarity) without (\ie sum-pooling) and with democratic re-weighting. 
\label{fig:relativeweights}}
\end{figure}

In the case of an arbitrary kernel $k(.,.)$, 
the assumptions required for convergence with Sinkhorn (matrix $\K$ nonnegative and fully indecomposable~\cite{K08}) are generally not satisfied.
Thus, a positive solution does not necessarily exist. 
Any optimisation algorithm may produce negative weights for kernels with negative values, 
which typically happen if $\sum_{\x'} k(x,x') < 0$. 
This is not desirable because it means that the weight computation is sensitive to new/deleted vectors in the set. 
We solve this problem by adopting the following pre-processing step:
we set all negative values to $0$ in $\K$.
The weights computed with this new matrix~$\K^+$ are positive with Sinkhorn's algorithm because all row/column sums are positive. 
The resulting kernel is not strictly a democratic kernel but tends towards more ``democracy", as illustrated by Figure~\ref{fig:relativeweights} for an arbitrary query image from Oxford5k dataset. 

\mypar{Computational cost.}
The cost of computing $\K$ is in $O(n^2D)$ and the cost of each Sinkhorn iteration is in $O(n^2)$.
For typical values of $D$, the first cost largely dominates.
Note that such a quadratic cost in $n$ might be too high for large values of $n$.

However, we can exploit the structure of certain embeddings.
Especially, the computation can be sped-up if the individual embeddings are block-sparse. 
By block-sparse we mean that the indices of the encoding can be partitioned into a set of groups where the activation
of one entry in a group means the activation of all entries in the group.
This is the case for instance of the VLAD and the SV where
each group of indices corresponds to a given cluster centroid.
This is also the case of the FV if we assume a hard assignment model
where each group corresponds to the gradients with respect to the parameters of a given Gaussian.
In such a case, the matrix $\K$ is block-diagonal.

Let us denote by $c$ the number of codewords in the VLAD/SV cases and the number of Gaussians in the FV case.
Using an inverted file type of structure, one can reduce the cost of computing $\K$ to $O(n^2D/c^2)$
by matching only the encodings that correspond to patches assigned to the same codeword/Gaussian\footnote{The $O(n^2D/c^2)$ 
complexity is based on the optimistic assumption that the same number of patches $n/c$ is assigned to each codeword.}.
Also, one can solve $c$ independent Sinkhorn algorithms block-by-block which reduces the cost of this step to $O(n^2/c)$.

\subsection{The BOW case}

We now show that our strategy is equivalent to the square-root component-wise normalisation in the case of bag-of-visual-words vectors 
(without inverse document frequency weighting) which is also known as the Hellinger kernel.

In the case of the BOW embedding (with hard assignment), $\bphi_\textrm{BOW}$ is defined by 
\begin{equation}
\bphi_\textrm{BOW}(\x)=[0,\dots,0, 1, 0, \dots,0]^\top. 
\label{equ:phibof}
\end{equation}
where the single $1$ corresponds to the index $i_\x$ of the closest centroid to descriptor $\x$.

These mapped vectors are summed up to produce the bag-of-visual-words vector 
\begin{equation}
\textrm{BOW}(\Xset)= [n_1(\Xset), \ldots, n_c(\Xset)]^\top,
\end{equation}
where $n_j(\Xset)$ (whose notation we simplify to $n_j$ in what follows)
is the number of descriptors assigned to visual word~$j$ in $\Xset$.

The match kernel matrix $\K_\textrm{BOW}$ is, 
up to a permutation (to assume that the vectors are ordered by increasing visual word indices), 
block diagonal with only 1 in the blocks: 
\begin{equation}
\K_\textrm{BOW}(\Xset,\Yset) = % \left[ 
\begin{array}{ccccll}
\framebox{\bf 1} & \framebox{\bf 0}  & \hdots & \framebox{\bf 0} & ~ & \updownarrow n_1 \\
\framebox{\bf 0} & \framebox{\bf 1}  & \hdots & \vdots           & ~ & \updownarrow n_2 \\
\vdots           & \vdots            & \ddots & \vdots           & ~ & \, \vdots \\
\framebox{\bf 0} & \hdots            & \hdots & \framebox{\bf 1} & ~ & \updownarrow n_c
\end{array}
\end{equation}

This matrix is positive, which means that the strategy to enforce only positive values has no effect. 
Similarly, the mapped vectors are already normalised to unit norm. 
A trivial solution to equation \eqref{eqn:dempb} is
\begin{equation}
\alpha_\Xset(\x)=1/\sqrt{n_{i_\x}}.
\end{equation} 
The resulting aggregated vector writes as
\begin{equation}
[\sqrt{n_1},\dots,\sqrt{n_i},\dots,\sqrt{n_c}]^\top.
\end{equation} 
This is the square-rooted version of the BOW which is known as the explicit feature map of the Hellinger kernel.
Interestingly, this component-wise normalisation is known to significantly improve the BOW for classification~\cite{boughorbel2005generalized,VZ10,PSL10}
and retrieval~\cite{JDS09a}. 
\medskip

Consider now the symmetric version of the Sinkhorn algorithm (Algorithm 1), where we set $\gamma=0.5$. 
The first iteration computes the sum of each row. 
If vector $\x$ is assigned to the visual word $i_\x$, 
then the sum is $n_j$ and $\alpha_\Xset(\x)=1/\sqrt{n_{i_\x}}$. 
In other terms, the algorithm reaches the fixed point in a single iteration.
For other values of $\gamma<0.5$, the algorithm also converges to this fixed point.

%% file: gmp.tex
% !TEX root = paper.tex

\section{Generalized max-pooling}
\label{sec:gmp}

While the democratic aggregation is equivalent to the Hellinger kernel in the BOW case,
we now seek a weighted aggregation mechanism that mimics the desirable properties of max-pooling for the BOW
and is extensible beyond count data -- hence the name Generalised Max Pooling (GMP).

One such property is the fact that the dot-product similarity between the max-pooled
representation which we denote $\bxi_{max}(\Xset)$
(or $\bxi_{max}$ for short when there is no ambiguity on the set $\x$)
and a single patch encoding $\bphi(\x)$ is a constant value.
To see this, let $c$ denote the codebook cardinality ($c=D$ in the BOW case)
and let $i_\x$ be the index of the closest codeword to patch $\x$.
As mentioned earlier, $\bphi(\x)$ is a binary vector with a single non-zero entry at index $i_\x$.
$\bxi_{max}$ is a binary representation where a 1 is indicative of the presence of the codeword in the image.
Consequently, we have:
\begin{equation}
  \label{eqn:maxprop}
  \bphi(\x)^\top \bxi_{max} = 1, \forall \x \in \Xset
\end{equation}
which means that $\bxi_{max}$ is equally similar to frequent and rare patches.
This occurs because frequent and rare patches 
contribute equally to the aggregated representation.

In the next section, we provide the GMP formulation, both in the primal and in the dual.
We then discuss the practical computation of the GMP representation
as well as its computational cost.
We finally show the relationship between the GMP and max-pooling in the BOW case.

\subsection{GMP Formulation}

We first introduce a primal formulation of the GMP which does not rely
on the explicit computation of a set of weights.
We then turn to the dual formulation that computes such weights explicitly.

\mypar{Primal formulation.}
Let $\bxi_{\gmp}(\Xset)$ (or $\bxi_{\gmp}$ for short when there is no ambiguity on the set $\x$)
denote the aggregated GMP representation for set $\Xset$.
We generalize the previous matching property \eqref{eqn:maxprop} to any embedding $\bphi(\x)$ and enforce
the dot-product similarity between each patch encoding $\bphi(\x)$ 
and the GMP representation $\bxi_{\gmp}$ to be a constant $C$:
\begin{equation} \label{eqn:match}
\bphi(\x)^\top \bxi_{\gmp} = C, \forall \x \in \Xset. 
\end{equation}
Note again that the value of the constant $C$ has no influence
as we typically $\ell_2$-normalize the final representation.
Therefore, we arbitrarily set this constant to $C=1$.
In matrix form, \eqref{eqn:match} can be rewritten as:
\begin{equation} \label{eqn:primal}
\bPhi^\top \bxi_{\gmp} = \vone_n.
\end{equation}
This is a linear system of $n$ equations with $D$ unknowns.
In general, this system might not have a solution (\eg when $D<n$)
or might have an infinite number of solutions (\eg when $D>n$).
Therefore, we turn \eqref{eqn:primal} into a least-squares regression problem and seek:
\begin{equation} \label{eqn:primal_ls}
\bxi_{\gmp} = \arg \min_{\bxi} ||\bPhi^\top \bxi - \vone_n||^2,
\end{equation}
with the additional constraint that $\bxi_{\gmp}$ has minimal norm in the case of an infinite number of solutions.
The previous problem admits a simple closed-form solution:
\begin{equation} \label{eqn:primalsol}
\bxi_{\gmp} = (\bPhi^\top)^+ \vone_n = (\bPhi \bPhi^\top)^+ \bPhi \vone_n
\end{equation}
where $^+$ denotes the pseudo-inverse and the second equality stems from the property
$\bM^+ = (\bM^\top \bM)^+ \bM^\top$ for any matrix $\bM$.

Since the pseudo-inverse is not a continuous operation 
it is beneficial to add a regularisation term to obtain a stable solution.
We introduce $\bxi_{\gmp,\lambda}$, the regularised GMP:
\begin{equation} \label{eqn:primalreg}
\bxi_{\gmp,\lambda} = \arg \min_{\bxi} ||\bPhi^\top \bxi - \vone_n||^2  + \lambda ||\bxi||^2.
\end{equation}
This is a ridge regression problem whose solution is:
\begin{equation} \label{eqn:primalsolreg}
\bxi_{\gmp,\lambda} = (\bPhi \bPhi^\top + \lambda \bI_D)^{-1} \bPhi \vone_n,
\end{equation}
where $\bI_D$ denotes the $D$-dimensional identity matrix.
The regularisation parameter $\lambda$ should be cross-validated.
For $\lambda$ very large, we have $\bxi_{\gmp,\lambda} \approx  \bPhi \vone_n / \lambda$
and we are back to sum pooling.
Therefore, $\lambda$ does not only play a regularisation role.
It also enables one to smoothly transition between the solution to \eqref{eqn:primalsol} ($\lambda=0$)
and sum pooling ($\lambda \rightarrow \infty$).

\mypar{Dual formulation.}
From \eqref{eqn:primalsolreg}, it is all but obvious that $\bxi_{\gmp}$
can be written as a weighted sum of the per-descriptor embeddings.
However, we note that the regularised GMP $\bxi_{\gmp,\lambda}$ is the solution to (\ref{eqn:primalreg})
and that, according to the representer theorem, 
$\bxi_{\gmp,\lambda}$ can be written as a linear combination of the embeddings:
\begin{equation} \label{eqn:rep}
\bxi_{\gmp,\lambda} = \bPhi \balpha_{\gmp,\lambda} 
\end{equation}
where $\balpha_{\gmp,\lambda}$ is the vector of weights.
By introducing $\bxi=\bPhi \balpha$ in the GMP objective (\ref{eqn:primalreg}),
we obtain: 
\begin{eqnarray}
\balpha_{\gmp,\lambda} & = & \arg \min_{\balpha} ||\bPhi^\top \bPhi \balpha - \vone_n||^2  + \lambda ||\bPhi \balpha||^2 \nonumber \\
                 & = & \arg \min_{\balpha} ||\K \balpha - \vone_n||^2  + \lambda \balpha^\top \K \balpha
\end{eqnarray}
which admits the following simple solution:
\begin{equation} \label{eqn:dualregsol}
\balpha_{\gmp,\lambda} = (\K+\lambda \bI_n)^{-1} \vone_n 
\end{equation}
where $\bI_n$ denotes the $n$-dimensional identity matrix.
Note that equation \eqref{eqn:dualregsol} only depends on the patch-to-patch similarity kernel $\K$,
not on the embeddings.

Once weights have been computed, the GMP representation is obtained by linearly re-weighting the
per-patch encodings -- see equation \eqref{eqn:rep}.
Note that in all our experiments we use the dual formulation to compute $\bxi_{\gmp,\lambda}$,
which is more efficient than the primal formulation given that typically $D > n$ for the descriptor sets we extract in our retrieval experiments.

\subsection{Computing the GMP in practice}
\label{sec:comp}

We now turn to the problem of computing $\bxi_{\gmp,\lambda}$.
We can solve Eq. \eqref{eqn:dualregsol} without explicitly inverting the matrix $(\K + \lambda \bI_n)$ 
(see Eq. \eqref{eqn:dualregsol}).
To do so, we use Conjugate Gradient Descent (CGD).
This might still be computationally intensive if the descriptor set cardinality $n$ 
is large and the matrix $\bPhi$ is full (cost in $O(n^2)$).

However, as was the case for the democratic aggregation,
we can take advantage of the special structure of certain embeddings.
Especially, the computation can be sped-up if the individual patch embeddings are block-sparse -- see section \ref{sec:demcomp}. 
In such a case, the matrix $\K$ is block-diagonal. 
Consequently $\K + \lambda \bI_n$ is block diagonal
and (\ref{eqn:dualregsol}) can be solved block-by-block (cost in $O(n^2/c)$).

\subsection{Relationship with max-pooling}

Recall that we denote by $\bphi(\Xset) = \{\bphi(\x_1), \ldots, \bphi(\x_n)\}$ the set of descriptor encodings of a given image.
We assume that the embeddings $\bphi(\x)$ are drawn from a finite codebook of possible embeddings, 
\ie $\bphi(x) \in \{q_1, \ldots, q_c\}$.
Note that the codewords $q_k$ might be binary or real-valued.
We denote by $\bQ$ the $D \times c$ codebook matrix of possible embeddings
where we recall that $D$ is the output encoding dimensionality
and $c$ is the codebook size.
We assume $\bQ = [q_1, \ldots, q_c]$ is orthonormal, \ie 
$\bQ^\top \bQ = \bI_c$ where $\bI_c$ is the $c \times c$ identity matrix.
For instance, in the case of the BOW with hard-coding,
$D=c$ and the $q_k$'s are binary with only the $k$-th 
entry equal to 1, so that $\bQ=\bI_c$.
We finally denote by $n_k$ the proportion of occurrences of $q_k$ in the set $\bphi(\Xset)$.

\mypar{Proposition.}
$\bxi_{\gmp}$ does not depend on the proportions $n_k$,
but only on the presence or absence of the $q_k$'s in $\bphi(\Xset)$.
\vspace{3mm}

\mypar{Proof.}
We denote by $\bN$ the $c \times c$ diagonal matrix that contains the values $n_1$, ..., $n_c$ on the diagonal. 
We rewrite $\bPhi \vone_n = \bQ \bN \vone_c$ and $\bPhi \bPhi^\top = \bQ \bN \bQ^\top$.
The latter quantity is the SVD decomposition of $\bPhi \bPhi^\top$ 
and therefore we have $(\bPhi \bPhi^\top)^+ = \bQ \bN^+ \bQ^\top$.
Hence equation (\ref{eqn:primalsol}) becomes $\bxi_{\gmp} = \bQ \bN^+ \bQ^\top \bQ \bN \vone_c = \bQ (\bN^+ \bN) \vone_c$.
Since $\bN$ is diagonal, its pseudo-inverse is diagonal and the values on the diagonal are equal to $1/n_k$ if
$n_k \neq 0$ and 0 if $n_k = 0$.
Therefore, $\bN^+ \bN$ is a diagonal matrix with element $k$ on the diagonal equal to 
$1$ if $n_k \neq 0$ and 0 otherwise.
Therefore we have
\begin{equation} \label{eqn:thm}
\bxi_{\gmp} = \sum_{k: n_k \neq 0} q_k,
\end{equation}
which does not depend on the proportions $n_k$,
just on the presence or absence of the $q_k$'s in $\bphi(\Xset)$.

\medskip

For the BOW hard-coding case, equation (\ref{eqn:thm}) shows that $\bxi_{\gmp}$ is 
a binary representation where each dimension informs on the presence/absence of each codeword in the image.
This is {\em exactly the max-pooled representation}.
Therefore, our pooling mechanism can be understood as a {\em generalisation of max-pooling}.

Note that there is no equivalence between the standard max-pooling and the GMP in the soft- or sparse-coding cases.
One benefit of the GMP however is that it is independent of a rotation of the encodings.
This is not the case of the standard max-pooling which operates on a per-dimension basis, but is the case for aggregation techniques such as the second- and third-order occurrence pooling proposed in \cite{koniusz2013comparison}.

%% file: exp_retrieval.tex
% !TEX root = paper.tex

\section{Experiments}
\label{sec:experiments}

This section presents results for our proposed aggregation methods.
We focus exclusively on instance-level image retrieval as we expect our methods to be most relevant for tasks that require retention of very localized information -- sometimes referred to as fine-grained tasks.
Instance-level retrieval is the ``ultimate'' fine-grained task as it requires the ability to distinguish object instances. 

Throughout this section, we only use the normalised
kernel $\Ksetn$ -- see Eq. \eqref{eqn:matchkernelnorm} -- meaning that the image vector is normalised to have unit Euclidean norm.
$\psis$ denotes the sum aggregation, $\psid$ the democratic aggregation and
$\psig$ the GMP aggregation.

\subsection{Datasets and evaluation protocol}

We adopt public datasets and corresponding evaluation protocols that are often used in the context of large scale image search.
All the unsupervised learning stages, \ie k-means clustering and PCA projection, are performed off-line using a distinct image collection
that contains neither the indexed database nor the query images.

\mypar{Oxford5k} \cite{PCISZ07} consists of 5,062 images of buildings and 55 query images corresponding to 11 distinct buildings in Oxford.
The search quality is measured by the mean average precision (mAP) computed over the 55 queries.
Images are annotated as either relevant, not relevant, or \emph{junk}, which indicates that it is unclear whether a user would consider the image as relevant or not.  Following the recommended protocol, the \emph{junk} images are removed from the ranking.
For the experiments on Oxford5k, all the learning stages are performed on the {\bf Paris6k} dataset~\cite{PCISZ08}.

\mypar{Oxford105k} is the combination of Oxford5k with
100k negative images, in order to evaluate the search quality on a large scale.

\mypar{INRIA Holidays}~\cite{JDS10a}. This dataset includes 1,491 photos of different locations and objects, 500 of them being used as queries. The search quality is measured by mAP, with the query removed from the ranked list.
For all three datasets, we repeat the experiments three times for our methods (for three distinct vocabularies) and report the mean performance.

\subsection{Implementation notes}
\label{sec:implementation}

\mypar{Local descriptors.}
They are extracted with the Hessian-affine detector~\cite{MiS04} and described by SIFT~\cite{Low04}.
We have used the same descriptors as provided in a previous paper~\cite{AZ13}.
We use the RootSIFT variant~\cite{AZ12} in all our experiments.

\mypar{Embedding functions $\bphi$.}
We consider two embedding functions in our experiments.
First is the Fisher Vector (FV) embedding which, given a local descriptor,
involves computing the gradient of its log-likelihood with respect to the parameters of a generative model
-- typically a Gaussian Mixture Model (GMM).
This was shown to be a competitive representation for instance-level image retrieval~\cite{PLSP10,JPDSPS12}.
We also consider the Triangulation embedding~\cite{JZ14} (T-embedding) which encodes the normalised residual of
the local descriptor with respect to a set of codewords.
It was recently shown to outperform the FV for the problem of instance-level image retrieval.
In what follows, we respectively denote by $\fvemb$ and $\triemb$ the FV and triangulation embedding functions.

\mypar{Power-law normalisation.}
As a common post-processing step~\cite{JDS09a,PSM10},
we apply power-law normalisation on the vector image representation,
and subsequently $\ell_2$-normalise it. This processing is parametrised by a constant $\alpha$ that controls the value of the exponent when modifying a component $a$ such that $a:=|a|^\alpha \textrm{sign}(a)$.
We use as standard $\alpha=0.5$ to ensure a fair comparison between the methods.
Note that we also include a specific analysis for this parameter.
While the goal of both the power-normalisation and the democratic/GMP aggregation
is to decrease the influence of bursty descriptors -- see Figure~\ref{fig:map_egs}
for a set of examples -- we have found that, in practice, their combination is beneficial.

\mypar{Rotation and Normalisation (RN).}
The power-law normalisation suppresses visual bursts, but
not the frequent co-occurrences that also corrupt the similarity
measure~\cite{CM10}. In VLAD, this problem is addressed~\cite{JC12} by
whitening the vectors. However, the whitening learning stage requires
a lot of input data and the smallest eigenvalues generate
artefacts. This makes such processing suitable only when producing very
short representations.
In \cite{koniusz2016}, power normalization is applied to the eigenvalues of the autocorrelation matrix of descriptors to reduce correlated bursts.
As an alternative~\cite{SQ13}, we apply
power-normalisation \emph{after} rotating the data with a PCA rotation
matrix learned on image vectors (from the learning set), i.e.\ no
whitening. This produces a similar effect to that of whitening, but
is more stable and not dependent on PCA eigenvalues. To avoid the full
eigen-decomposition and the need to use too many images for the
learning stage, we compute the first 1,000 eigenvectors and apply
Gram-Schmid orthogonalisation on the reminder of the space (orthogonal complement to these first eigenvectors) to produce
a complete basis.  After this rotation, we apply the regular power-law
normalisation, which then jointly addresses the bursts and
co-occurrences by selecting a basis capturing both phenomenons on the
first components.

\subsection{Impact of the methods and parameters}

\begin{figure}
\centering
\includegraphics[height=0.7\linewidth]{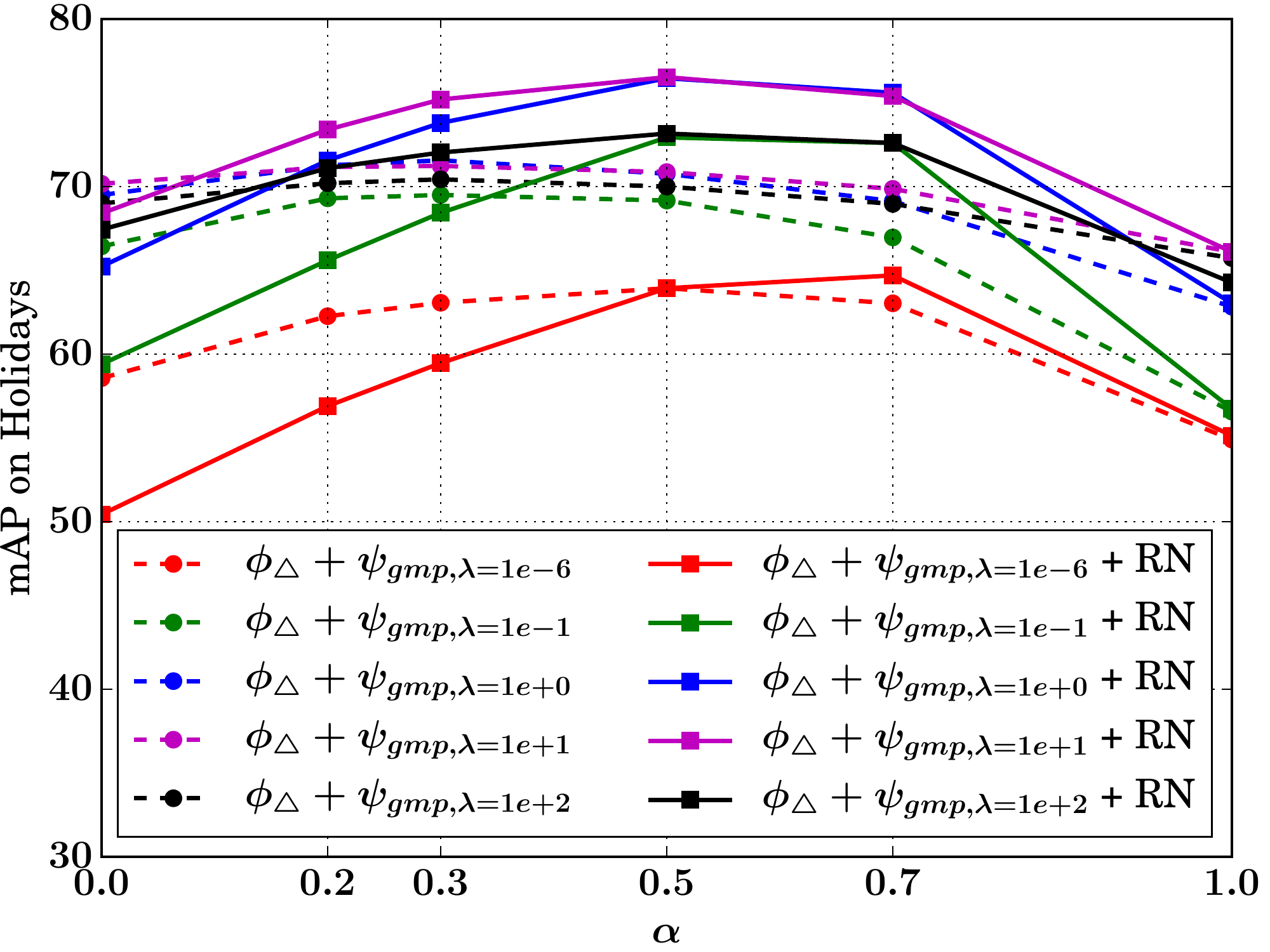}\\
\includegraphics[height=0.7\linewidth]{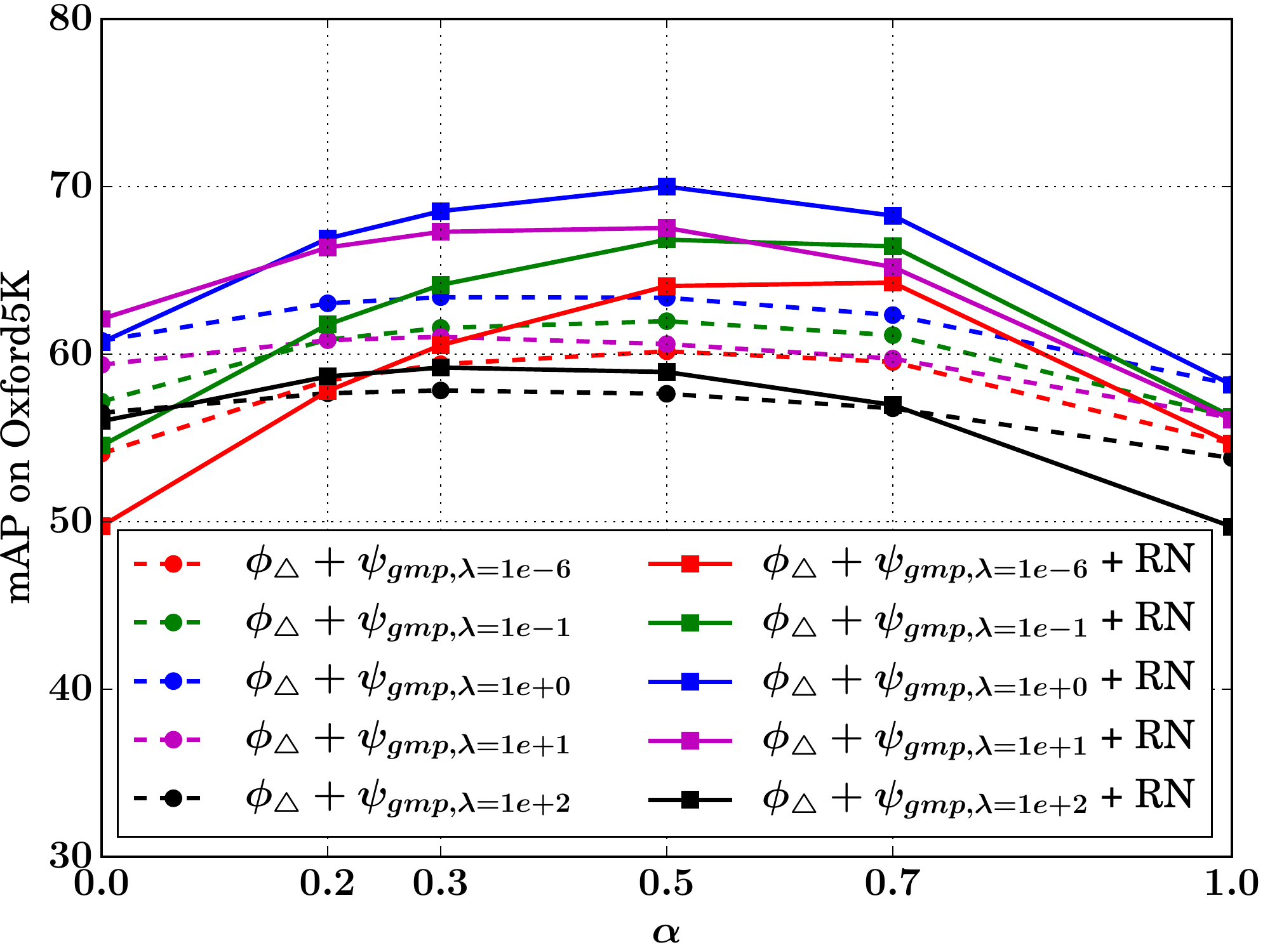}

\caption{Impact of the GMP parameter $\lambda$ on performance for the \embname $\triemb$.
  Top: Holidays. Bottom: Oxford5k.
  Results are shown both with and without Rotation and Normalization (RN).
  mAP is reported as a function of the power-law normalisation exponent $\alpha$. $\k=64$.
Note, $\alpha=0$ amounts to binarising the vector.
\label{fig:impact_lambda}}
\vspace{-5pt}
\end{figure}

\begin{figure*}[!ht]
\centering

\includegraphics[height=0.23\textwidth]{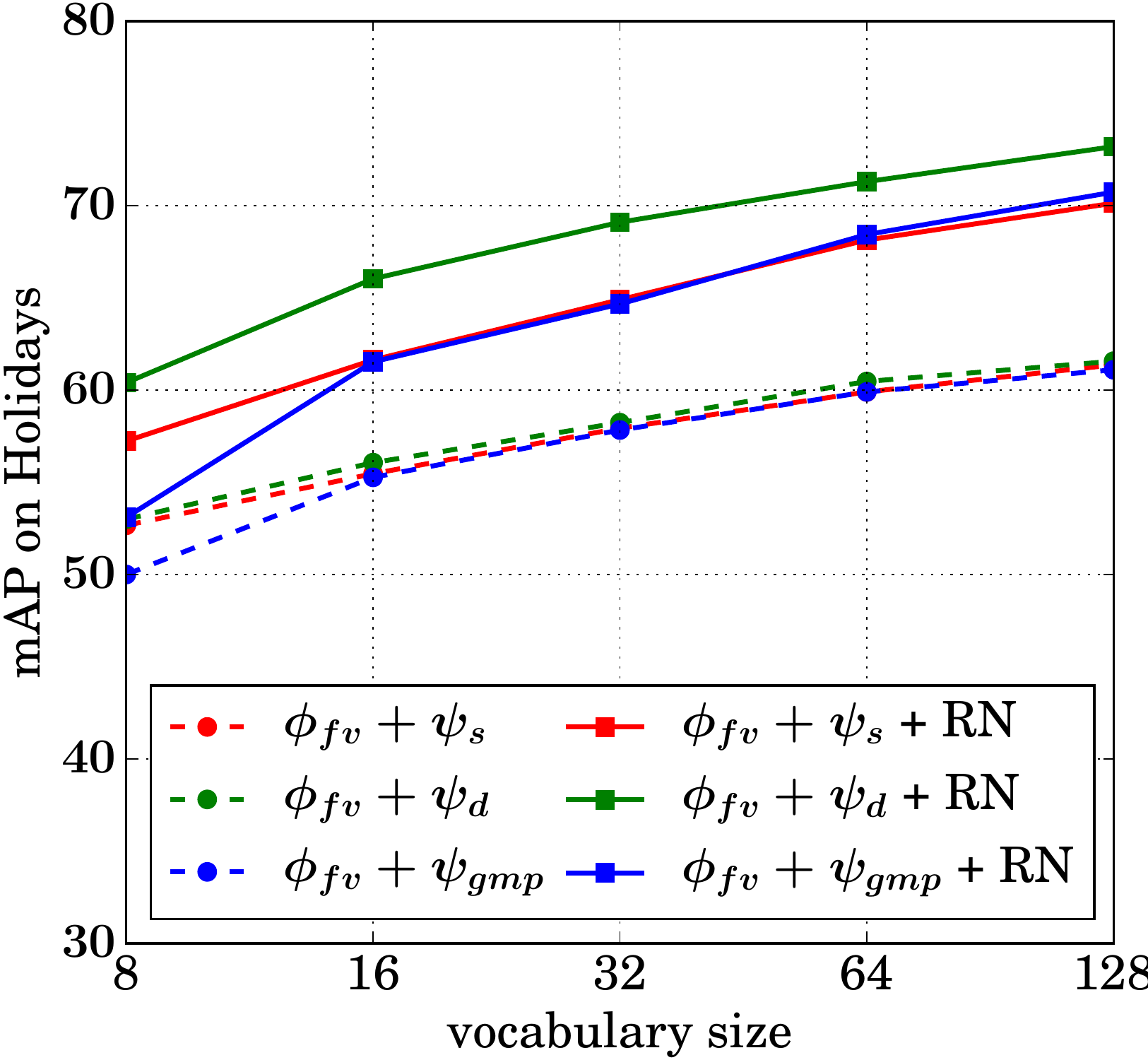}%
\includegraphics[height=0.23\textwidth]{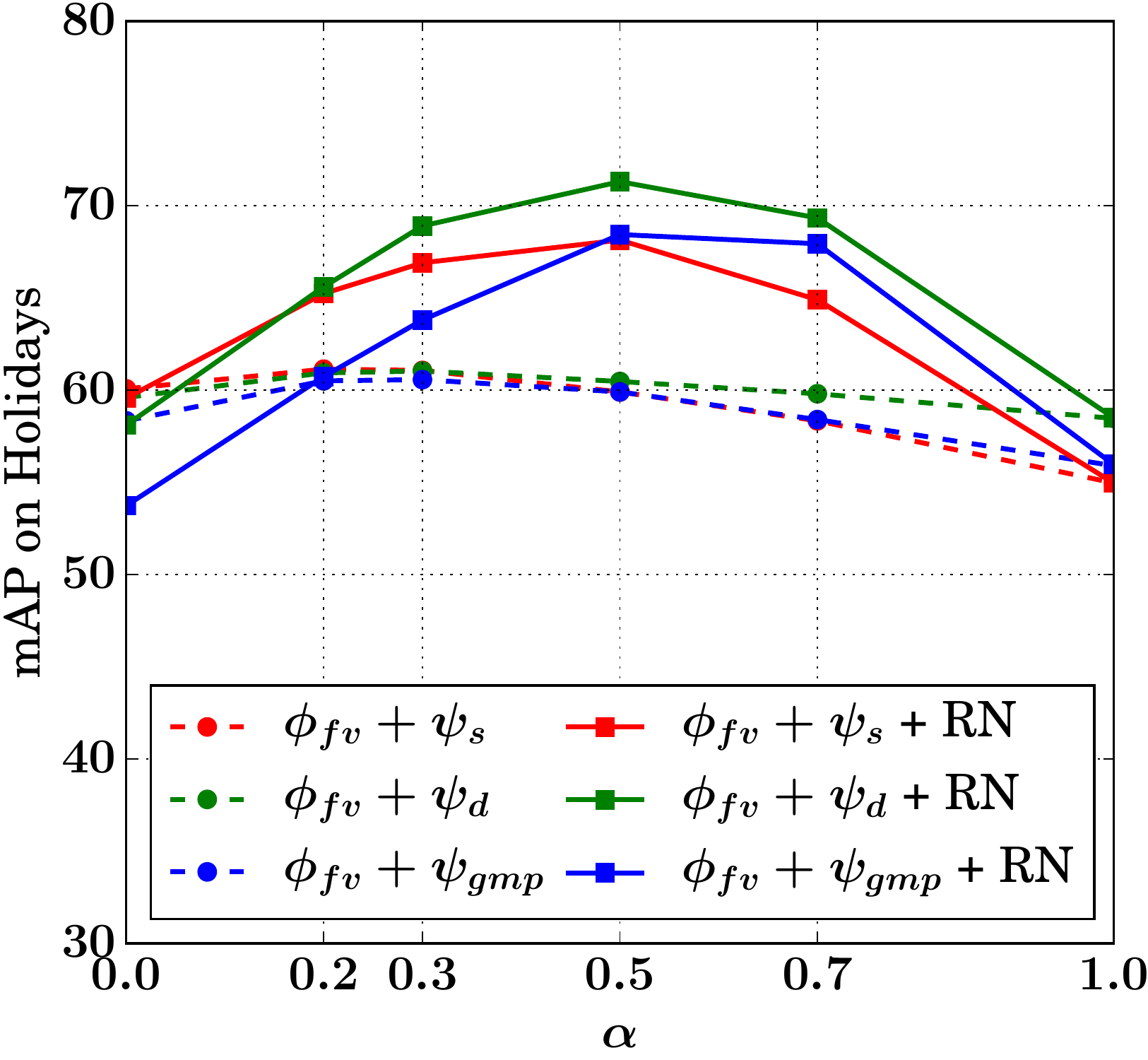}%
\includegraphics[height=0.23\textwidth]{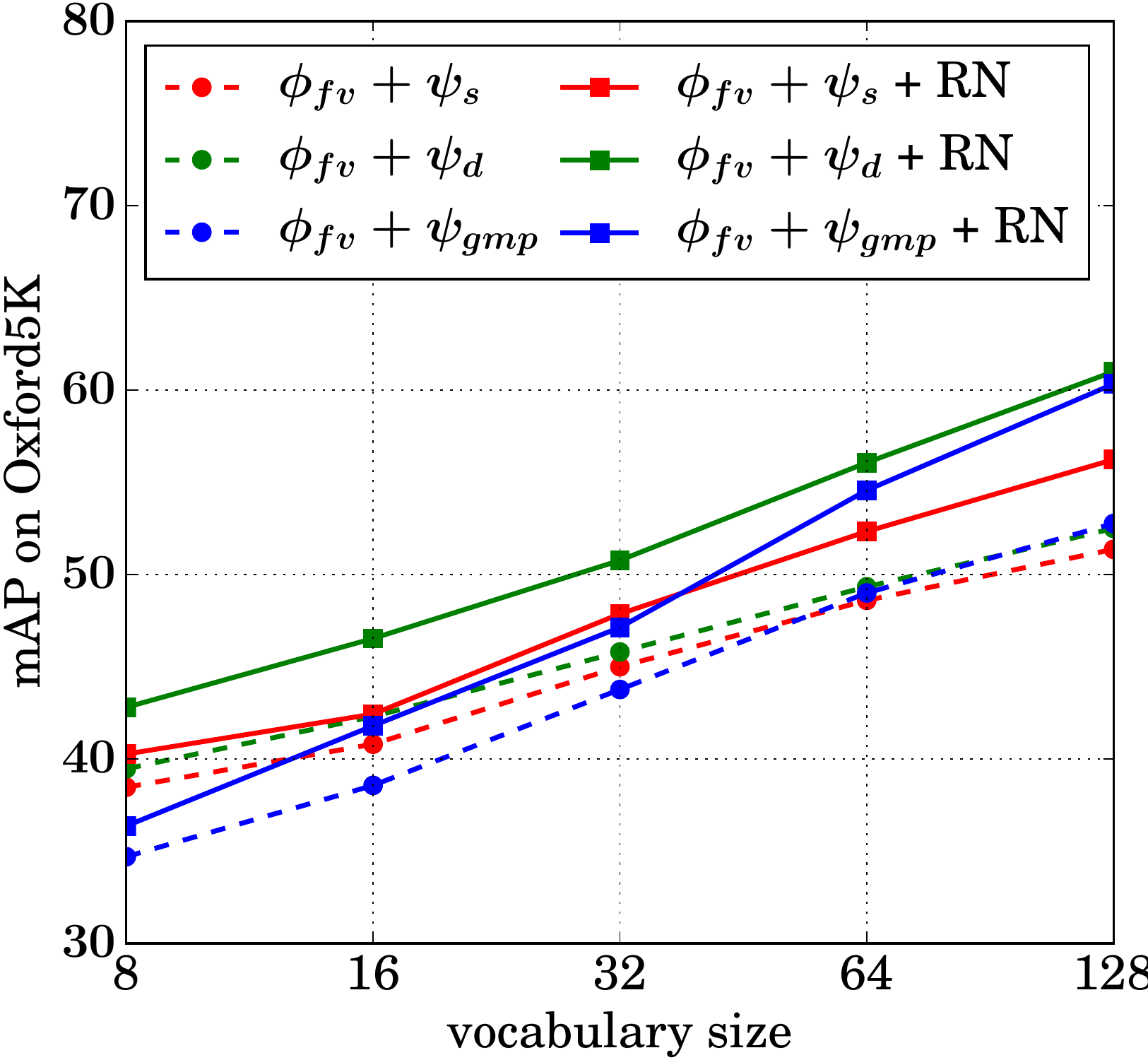}%
\includegraphics[height=0.23\textwidth]{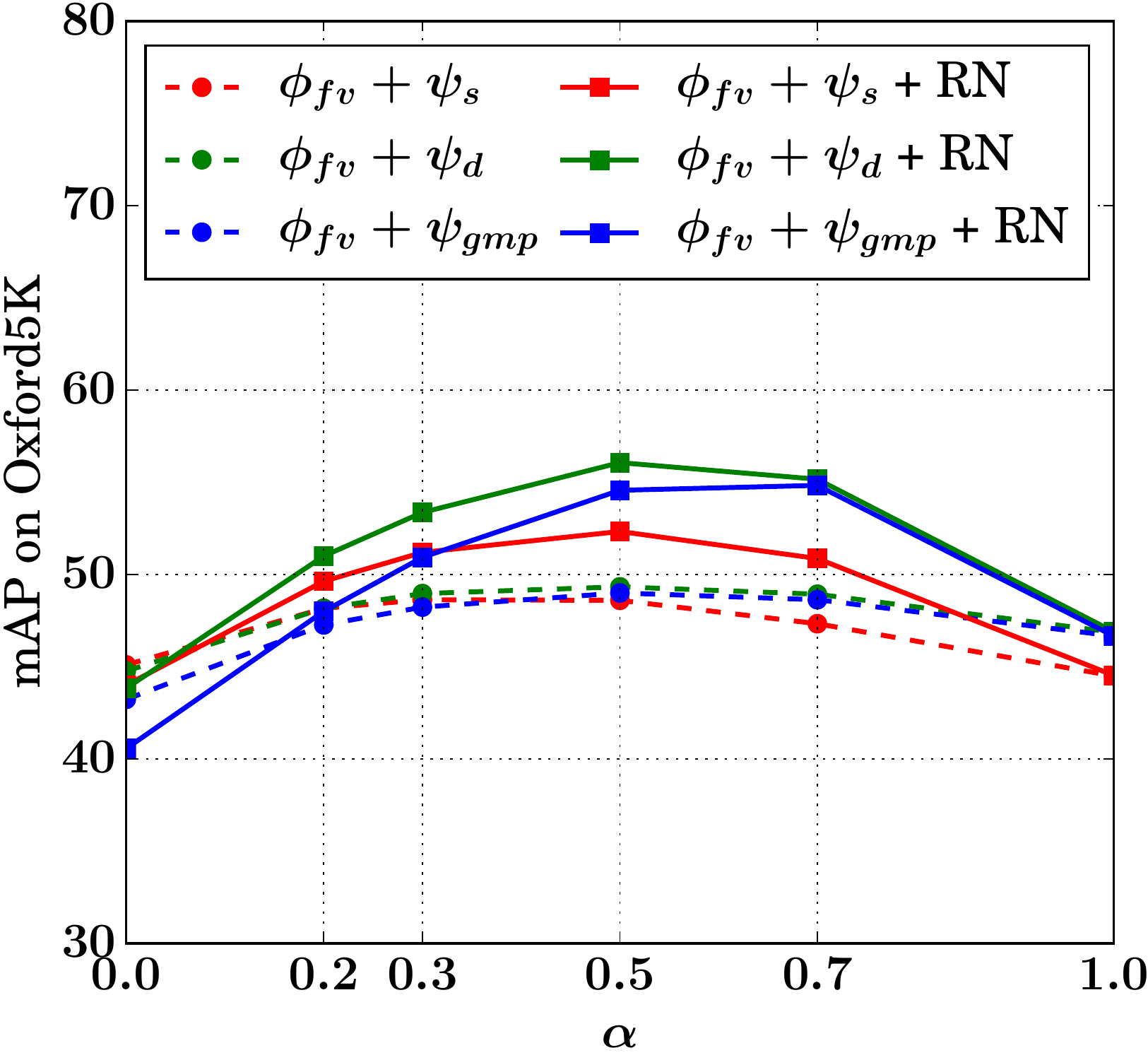}

\includegraphics[height=0.23\textwidth]{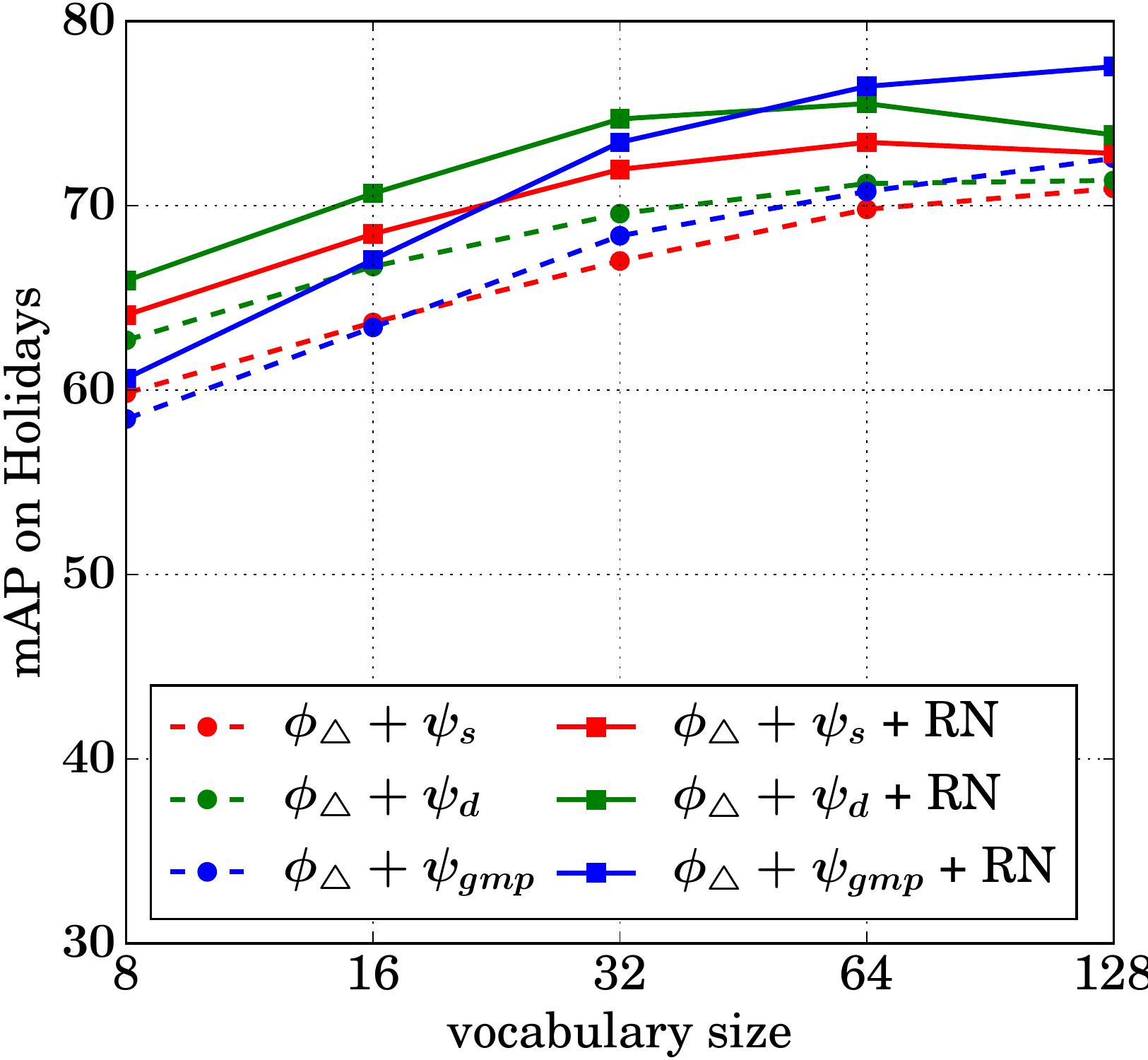}%
\includegraphics[height=0.23\textwidth]{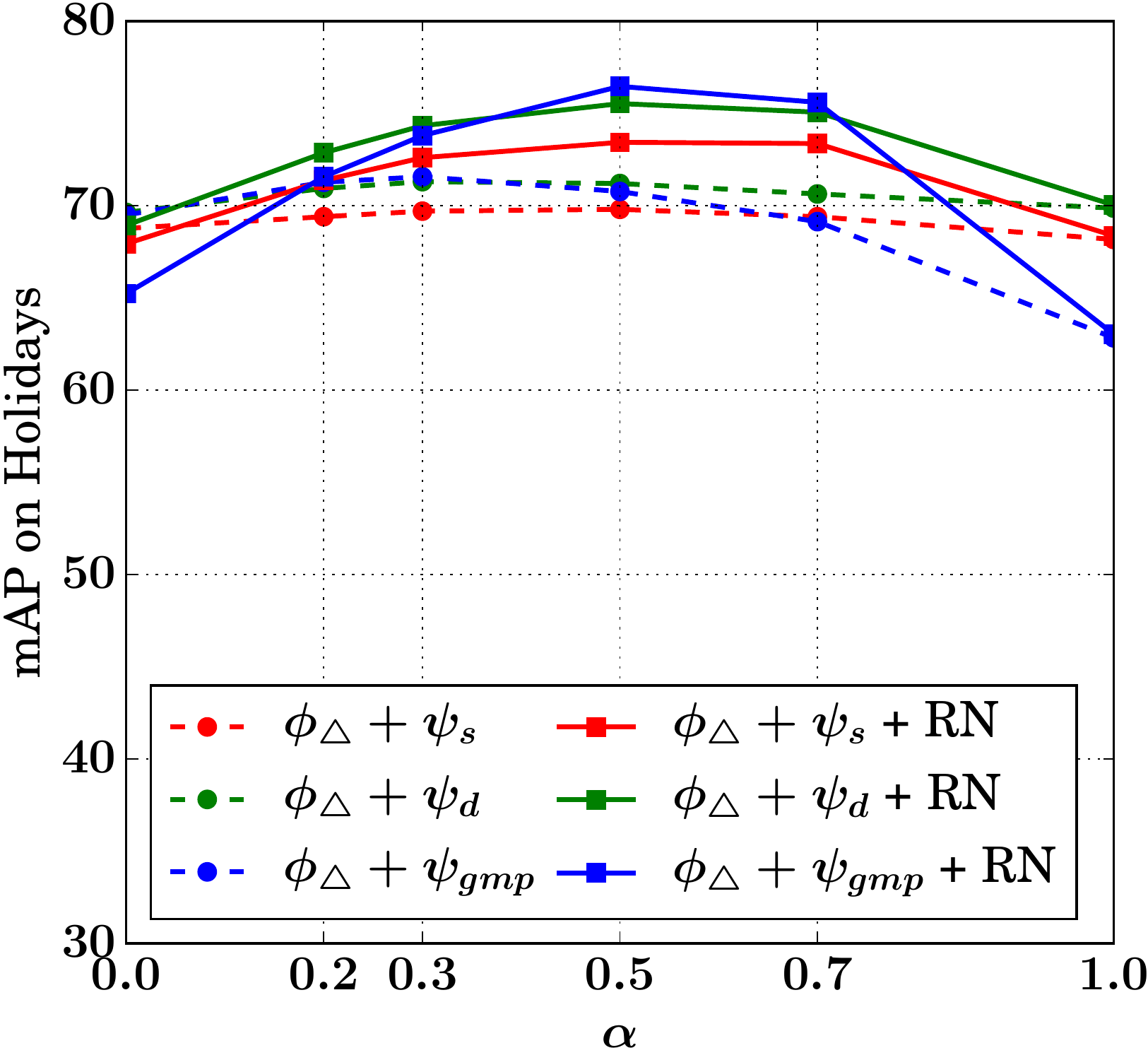}%
\includegraphics[height=0.23\textwidth]{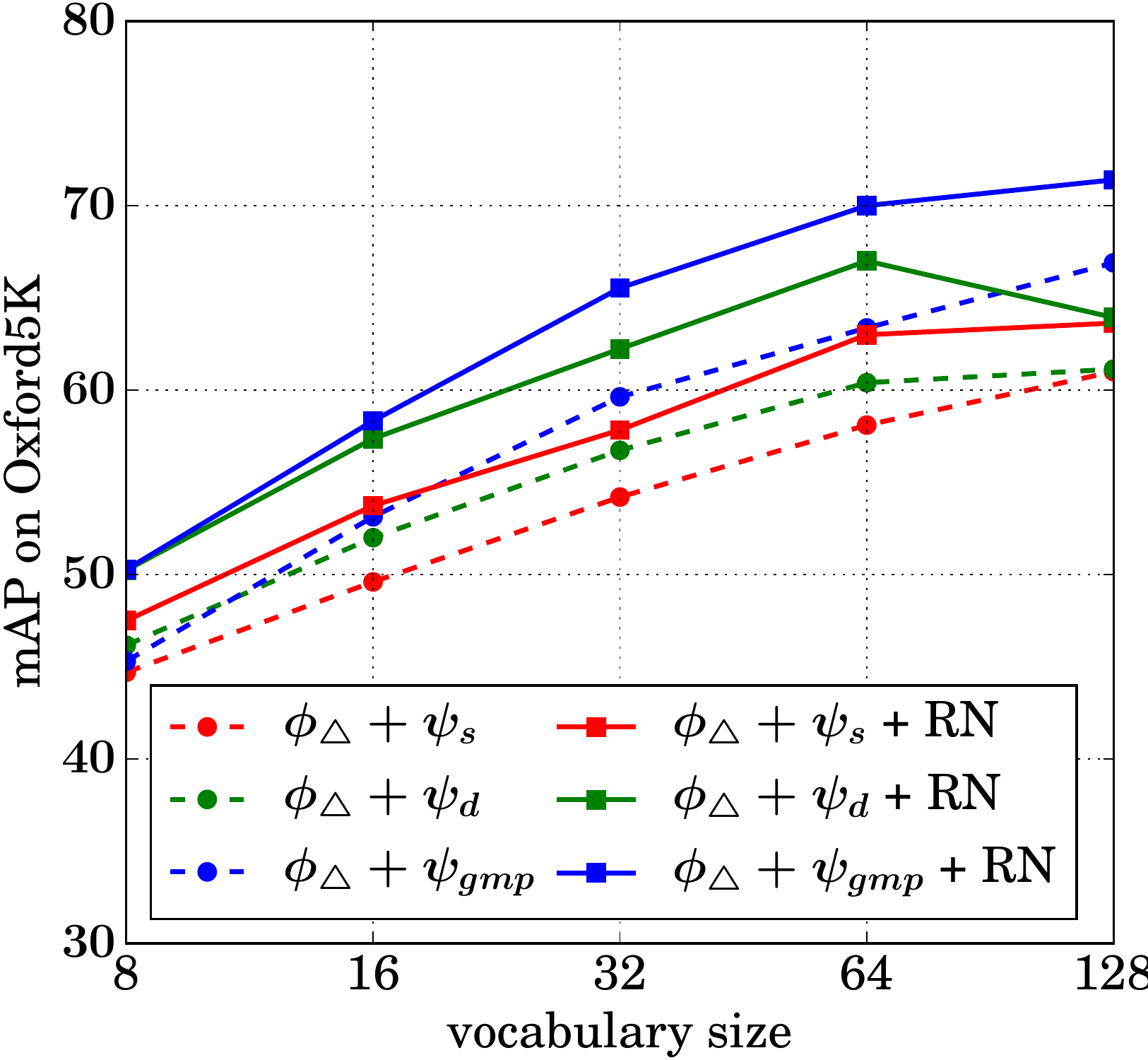}%
\includegraphics[height=0.23\textwidth]{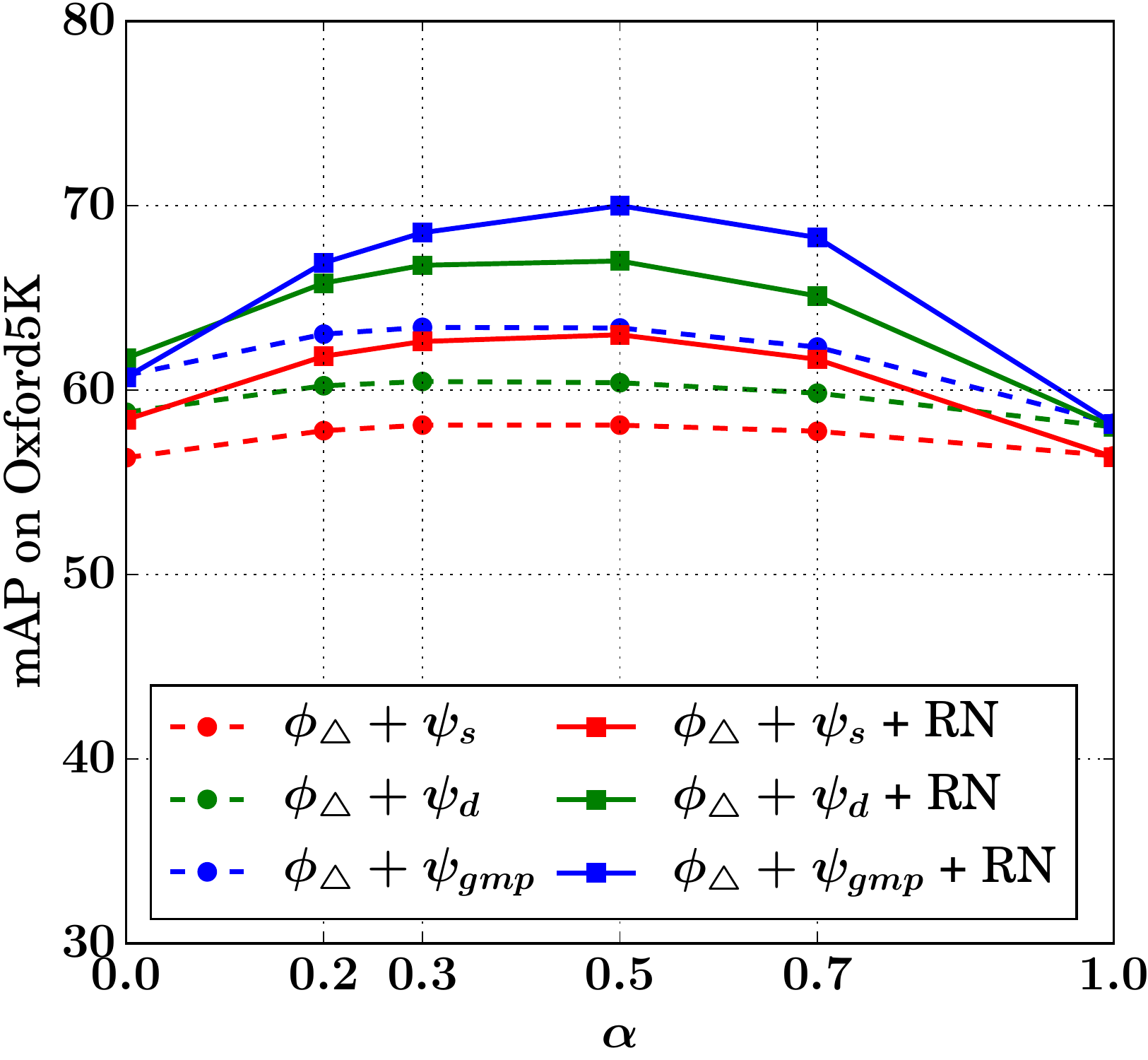}%

\caption{Impact of the parameters on performance for different embeddings and different aggregation methods: results for FV embeddings are on the top row, while those for \embname $\triemb$ are on the bottom row. Results are shown (both with and without RN) for sum aggregation $\psis$, democratic aggregation~$\psid$ and GMP~$\psig$. mAP is reported as a function of vocabulary size $\k$, and of the power-law normalisation exponent $\alpha$ (with $\k=64$ in this case).
Note, $\alpha=0$ amounts to binarising the vector.
\label{fig:impact_params}}
\vspace{-5pt}
\end{figure*}

Democratic aggregation introduces no extra parameter compared with existing
techniques, apart from constants that do not have a noticeable impact on the effectiveness of the method, like the number of iterations in Sinkhorn.
GMP includes one new parameter, the regularisation factor $\lambda$.
Setting $\lambda=1$ was found to perform well in a wide range of experiments, as shown in Figure~\ref{fig:impact_lambda}.
Therefore we set $\lambda=1$ in all remaining experiments.
The main parameters are therefore the
vocabulary size $\k$ and the parameter $\alpha$ associated with
power-law normalisation. The analysis of these two parameters is shown in Figure~\ref{fig:impact_params} for Holidays and Oxford5k.
To complement these curves, Table~\ref{tab:impact_method} shows the impact of our methods step by
step for a fixed vocabulary size on Oxford5k, Oxford105k and Holidays.
The conclusions we draw are similar on both datasets and are as follows.

\mypar{Vocabulary size.}
For all representations, including democratic aggregation and GMP,
the performance is an increasing function of the vocabulary size.
The improvement due to the aggregation mechanism tends to be smaller for larger vocabularies.
This is expected, as for larger vocabularies the interaction between the descriptors is less important than for small ones. For $\k > 128$, the benefits of our aggregation strategies are not worth the computational overhead, especially for dense embeddings such as the T-embedding.

\mypar{Our aggregation strategies} give a significant boost in performance.
As is expected, it improves the performance when no power-law is applied.
Moreover, the analysis of the power parameter
$\alpha$
also reveals that
our aggregation methods $\psid$ and $\psig$ are complementary to the power-law normalisation,
as using them both improves the score.

\mypar{Power-law normalisation and RN.}
Combining power-law normalisation with our methods is more beneficial (the curves $\triemb+\psis$ and $\triemb+\psid$ are less flat) when we also employ RN.
In particular, this normalisation gives a large improvement in performance when used with the (standard) parameter $\alpha=0.5$.
In summary, the best results are obtained with the system that combines
T-embedding as the per-patch embedding,
GMP as the aggregation mechanism and power-law normalisation with RN for the normalisation.

\mypar{Dimensionality reduction.} In order to get shorter representations, we keep the first $D'$ components, after RN normalisation, of the vector produced by our embeddings. Table~\ref{tab:impact_method} reports the performance for short vectors of varying dimensionality, $D'=128$ to $1024$. Dimensionality reduction induces a loss in performance that is comparatively larger for our method than with representations derived from convolutional neural networks (CNNs), which concurs with the observation of Babenko \etal~\cite{babenko2014neural}. This suggests that, when combined with Fisher or Triangulation embedding, the operating points on which our aggregation method is useful are mid-size representations. In this case we achieve state-of-the-art performance on Oxford5k and Oxford105k.

 \newcommand\mstd[1]{\scriptsize \,$\pm$\it#1}

\begin{table}[t]
\caption{Impact of our methods on the performance. First we evaluate the Fisher vector and combine it with $\psid$ and $\psig$. Then we consider  \embname $\triemb$ with sum ($\psis$), democratic ($\psid$) and GMP ($\psig$) aggregation, and show the boost given by RN. Finally, we present results after dimensionality reduction to short vectors.
The evaluation is carried out on the following standard benchmarks: Holidays, Oxford5k and Oxford105k.
$\k=64$. \smallskip
\label{tab:impact_method}}
\centering
{\scriptsize
\begin{tabular}{|l|l|l@{\hspace{10pt}}l@{\hspace{8pt}}c@{\hspace{5pt}}|}
\hline
                         & dim. red. &  \multicolumn{3}{c|}{mAP} \\
method $\downarrow$      & to $\rightarrow D'$          & Holidays & Oxford5k & Ox105k  \\
\hline

$\fvemb$ + $\psis$ & \hspace{12pt} -- & 61.7\mstd{0.2} & 49.5\mstd{1.8} & 43.3\mstd{1.3} \\
$\fvemb$ + $\psid$ & \hspace{12pt} -- & 62.1\mstd{0.2} & 50.7\mstd{2.0} & 44.6\mstd{1.6} \\
$\fvemb$ + $\psig$ & \hspace{12pt} -- & 61.4\mstd{0.2} & 51.1\mstd{1.1} & 44.7\mstd{0.8} \\

$\fvemb$ + $\psis$ + RN & \hspace{12pt} -- & 68.7\mstd{0.0} & 53.4\mstd{1.5} & 44.3\mstd{1.9} \\
$\fvemb$ + $\psid$ + RN & \hspace{12pt} -- & 72.5\mstd{0.7} & 56.0\mstd{2.1}& 47.6\mstd{1.4} \\
$\fvemb$ + $\psig$ + RN & \hspace{12pt} -- & 69.4\mstd{1.3} & 54.6\mstd{2.0} & 45.9\mstd{1.4} \\

$\triemb$ + $\psis$      & \hspace{12pt}   --       & 69.8\mstd{0.6} & 58.1\mstd{0.3} & 51.3\mstd{0.3} \\
$\triemb$ + $\psid$      & \hspace{12pt}   --       & 71.2\mstd{0.1} & 60.4\mstd{0.1} & 54.3\mstd{0.1} \\
$\triemb$ + $\psig$      & \hspace{12pt}   --       & 70.8\mstd{0.1} & 63.4\mstd{0.0} & 57.9\mstd{0.0} \\

$\triemb$ + $\psis$ + RN & \hspace{12pt}   --       & 73.4\mstd{0.0} & 63.0\mstd{0.2} & 55.0\mstd{0.1} \\
$\triemb$ + $\psid$ + RN & \hspace{12pt}   --       & 75.5\mstd{0.1} & 67.0\mstd{0.1} & 60.2\mstd{0.1} \\
$\triemb$ + $\psig$ + RN & \hspace{12pt}   --       & {\bf 76.5}\mstd{0.3} & {\bf 70.0}\mstd{0.1} & {\bf 64.4}\mstd{0.1} \\

\hline
$\triemb$ + $\psid$ + RN & $\rightarrow$ 1,024     & 69.4\mstd{0.2} & 55.1\mstd{0.4} & 49.3\mstd{0.3} \\
$\triemb$ + $\psid$ + RN & $\rightarrow$ 512       & 65.6\mstd{0.6} & 50.9\mstd{0.6} & 45.3\mstd{0.2} \\
$\triemb$ + $\psid$ + RN & $\rightarrow$ 256       & 59.2\mstd{0.6} & 45.3\mstd{1.2} & 39.2\mstd{1.0} \\
$\triemb$ + $\psid$ + RN & $\rightarrow$ 128       & 54.4\mstd{0.6} & 38.9\mstd{0.3} & 33.0\mstd{0.4} \\
\hline

$\triemb$ + $\psig$ + RN & $\rightarrow$ 1,024     & 71.6\mstd{0.3} & 58.3\mstd{0.3} & 52.5\mstd{0.1} \\
$\triemb$ + $\psig$ + RN & $\rightarrow$ 512       & 68.9\mstd{0.1} & 53.8\mstd{0.4} & 48.4\mstd{0.2} \\
$\triemb$ + $\psig$ + RN & $\rightarrow$ 256       & 66.0\mstd{0.2} & 48.5\mstd{0.2} & 42.7\mstd{0.0} \\
$\triemb$ + $\psig$ + RN & $\rightarrow$ 128       & 62.4\mstd{0.9} & 42.2\mstd{0.1} & 36.6\mstd{0.1} \\
\hline
\end{tabular}}
\end{table}

\subsection{Comparison with the state-of-the-art}

\mypar{Comparison with related baselines.}
We consider as baselines recent works
targeting the same application scenario and similar representations,
\ie that represent an image by a vector that may be subsequently
reduced~\cite{JPDSPS12}. We compare with works recently
published on similar mid-size vector
representations~\cite{AZ13,JPDSPS12}. We also compare with our
re-implemented (improved) version of VLAD and Fisher vectors that
integrates RootSIFT. This baseline, by itself, approaches or
outperforms the state of the art by combining most of the
effective ingredients.

Table~\ref{tab:stateoftheart} shows that our method outperforms the compared methods by a large margin on all datasets. The gain over a recent paper~\cite{AZ13} using a larger vocabulary is
{\bf +11.2\%} in mAP on Holidays and
{\bf +14.2\%}
in mAP on Oxford5k. Compared with our improved Fisher baseline using the same vocabulary size (see first row of Table~\ref{tab:impact_method}), the gain is
{\bf +14.8\%}
in mAP for Holidays,
{\bf +20.5\%}
for Oxford5k and
{\bf +21.1\%}
for Oxford105k.
Even when reducing the dimensionality to $D'=1,024$ components, we outperform all other similar methods by a large margin, with a much smaller vector representation. Only when reducing the vector to $D'=128$ components does our method give on average slightly lower results than those reported by Arandjelovi\'c and Zisserman~\cite{AZ13}.

\begin{table}
\caption{Comparison with the state of the art for short and intermediate representations
produced from the same descriptors.
The last two rows show the performance after reducing our vector from 8,064 to 1,024 or 128 components. %\smallskip
\label{tab:stateoftheart}}
\centering{\scriptsize
\begin{tabular}{|l@{\hspace{5pt}}rr|cc@{\hspace{8pt}}c@{\hspace{5pt}}|}
\hline
                         & $\k$   & $\D$ \ \ \   & \multicolumn{3}{c|}{mAP} \\
method $\downarrow$      &        &        & 
Holidays & Ox5k & Ox105k \\
\hline
BOW~\cite{JPDSPS12}      & 20k    & 20,000 & 43.7 & 35.4 & --\ \  \\
BOW~\cite{JPDSPS12}      & 200k   & 200,000& 54.0 & 36.4 & --\ \  \\
VLAD~\cite{JPDSPS12}     & 64     &  4,096 & 55.6 & 37.8 & --\ \  \\
Fisher~\cite{JPDSPS12}   & 64     &  4,096 & 59.5 & 41.8 & --\ \  \\
VLAD-intra~\cite{AZ13}   & 256    & 32,536 & 65.3 & 55.8 & --\ \  \\
VLAD-intra~\cite{AZ13}   & 256    & $\rightarrow$ 128 & 62.5 & 44.8 & 37.4 \ \  \\
\hline
\multicolumn{6}{|c|}{\it Our methods} \\
\hline
$\triemb$ + $\psis$ + RN & 16  & 1,920  & 68.5 & 53.7 & 46.2 \\
$\triemb$ + $\psis$ + RN & 64  & 8,064  & 73.4 & 63.0 & 55.0 \\
\hline
$\triemb$ + $\psid$ + RN & 16  & 1,920  & 70.7 & 57.4 & 50.7 \\
$\triemb$ + $\psid$ + RN & 64  & 8,064  & 75.5 & 67.0 & 60.2 \\
\hline
$\triemb$ + $\psig$ + RN & 16  & 1,920  & 67.1 & 58.3 & 51.4 \\
$\triemb$ + $\psig$ + RN & 64  & 8,064  & {\bf 76.5} & {\bf 70.0} & {\bf 64.4} \\
\hline
$\triemb$ + $\psid$ + RN &  16  & $\rightarrow$ 128    & 60.7\mstd{0.0} & 42.6\mstd{1.2} & 35.4\mstd{0.6} \\
$\triemb$ + $\psid$ + RN &  64  & $\rightarrow$ 1,024  & 69.4\mstd{0.2} & 55.1\mstd{0.4} & 49.3\mstd{0.3} \\
\hline
$\triemb$ + $\psig$ + RN &  16  & $\rightarrow$ 128    & 61.5\mstd{0.6} & 44.4\mstd{0.0} & 37.4\mstd{0.0} \\
$\triemb$ + $\psig$ + RN &  64  & $\rightarrow$ 1,024  & 71.6\mstd{0.3} & 58.3\mstd{0.3} & 52.5\mstd{0.1} \\
\hline
\end{tabular}}%}
\end{table}

\mypar{Comparison with deep baselines.}
Recently, it has been proposed to use CNNs~\cite{lecun1998}
as feature extractors for instance-level image retrieval~\cite{razavian2014cnn,GWGL14,babenko2014neural,azizpour2015factors,ng2015exploiting,babenko2015aggregating}.
In a nutshell, this approach involves training a convnet on a large dataset
of labeled images such as ImageNet~\cite{deng2009imagenet} and then,
given an image, to use as image feature(s) the output of intermediate layers as computed during the forward pass.

Two main variations exist around this same principle.
The first approach involves extracting a single CNN representation per image,
usually the output of the penultimate layer~\cite{razavian2014cnn,babenko2014neural,azizpour2015factors}.
These features are subsequently compared using the cosine similarity or Euclidean distance.
The second approach involves extracting multiple patch-level CNN representations per image.
These local CNN descriptors are then treated as local descriptors as is the case of the SIFT.
They can then be embedded using VLAD encoding, FV encoding or T-embedding~\cite{GWGL14,ng2015exploiting,babenko2015aggregating}
and are typically aggregated with sum-pooling and subsequently normalised.
They can also be aggregated as is without further encoding~\cite{babenko2015aggregating}.

Table~\ref{tab:deepbaseline} compares our results with CNN-based results.
We underline that not all of these results are directly comparable to ours -- see the caption for details.
Surprisingly, while still relying on weaker SIFT features
and while not leveraging gargantuan amounts of external data,
our results are competitive on Oxford5k and Oxford105k.
Our results are still state-of-the-art for the most costly operating points,
while CNN features perform best for very short image representations,
because they suffer less from dimensionality reduction~\cite{babenko2014neural}.
We note that our own results could be improved, either by using supervised dimensionality reduction
techniques as proposed in~\cite{gordo2012leveraging},
or by using strong CNN local features (as done in~\cite{GWGL14,ng2015exploiting,babenko2015aggregating})
as opposed to SIFT features.
Hence, we do not see our democratic and GMP aggregation mechanisms as in competition with works leveraging CNNs.
On the contrary, they could be combined to obtain better results\footnote{
\cite{babenko2015aggregating} also mentions the possibility but does not report results.}.

\begin{table}
\caption{Comparison with baselines produced using CNN features.
Note that the asterisk $^*$ indicates results obtained without following the standard experimental protocols
(which we fully follow in our own experiments).
Such non standard practices involve on Holidays manually rotating images and on Oxford querying with larger regions than those provided.
\label{tab:deepbaseline}}
%{\scriptsize
\centering{\scriptsize
\begin{tabular}{|l@{\hspace{5pt}}rr|cc@{\hspace{8pt}}c@{\hspace{5pt}}|}
\hline
                         & $\k$   & $\D$ \ \ \   & \multicolumn{3}{c|}{mAP} \\
method $\downarrow$      &        &        & 
Holidays & Ox5k & Ox105k \\
\hline
\multicolumn{6}{|c|}{\it Our methods} \\
\hline
$\triemb$ + $\psig$ + RN & 64   & 8,064  & 76.5 & {\bf 70.0} & {\bf 64.4} \\
$\triemb$ + $\psig$ + RN &  64  & $\rightarrow$ 1,024  & 71.6 & 58.3 & 52.5 \\
$\triemb$ + $\psig$ + RN &  16  & $\rightarrow$ 128    & 61.5 & 44.4 & 37.4 \\
\hline
\multicolumn{6}{|c|}{\it One CNN feature extracted per image} \\
\hline
Babenko \etal~\cite{babenko2014neural}  & - & 4,096  & 74.9$^*$ & 43.5 & 39.2 \\
                                  & - & $\rightarrow$  512 & 74.9$^*$ & 43.5  & 39.2 \\
                                  & - & $\rightarrow$  128 & 74.7$^*$ & 43.3  & 38.8 \\
Retraining on landmarks  & - & 4,096  &  79.3$^*$ & 54.5  & 51.2 \\
                        & - & $\rightarrow$ 512 & 78.9$^*$  & 55.7  & 52.2 \\
                        & - & $\rightarrow$ 128 & 78.9$^*$  & 55.7  & 52.3 \\
Razavian~\etal~\cite{RSMC15}   & - & $\rightarrow$ 256 & 71.6$^*$  & 53.3  & 48.9 \\
\hline
\multicolumn{6}{|c|}{\it Multiple CNN features extracted + aggregated} \\
\hline
Gong~\etal~\cite{GWGL14} & 100 & 12,288 & 78.8 & - & - \\
                         & 100 & $\rightarrow$ 2,048 & 80.8 & - & - \\
Ng~\etal~\cite{ng2015exploiting} & 100 & varying & \bf{84.0} & 64.9$^*$ & - \\
                                 & 100 & $\rightarrow$ 128 &  83.6 & 59.3$^*$ &- \\
Babenko~\cite{babenko2015aggregating} & - & $\rightarrow$ 256 & 80.2$^*$ & 58.9$^*$ & 57.8$^*$ \\
cropped queries  & - & $\rightarrow$ 256 & - & 53.1 & 50.1 \\
\hline
\end{tabular}}%}
\end{table}

\mypar{Comparison with the best reported results.}
We now compare in Table~\ref{tab:costbaseline} our results with the very best results obtained in the literature on the considered datasets.
All these pipelines involve the extraction and matching of multiple regions per image.
Hence, these methods do not represent an image with a single descriptor as is the case of our approach.
Consequently, they are significantly more costly than the proposed approach, whether in terms of memory or computational usage.

\begin{table}
\caption{Comparison with costly baselines that involve the matching of multiple regions per image.
As is the case in Table~\ref{tab:deepbaseline}, the asterisk $^*$ indicates that the results were obtained
using a non-standard experimental protocol.
\label{tab:costbaseline}}
%{\scriptsize
\centering{\scriptsize
\begin{tabular}{|l@{\hspace{5pt}}rr|cc@{\hspace{8pt}}c@{\hspace{5pt}}|}
\hline
                         & $\k$   & $\D$ \ \ \   & \multicolumn{3}{c|}{mAP} \\
method $\downarrow$      &        &        & 
Holidays & Ox5k & Ox105k \\
\hline
\multicolumn{6}{|c|}{\it Best reported results} \\
\hline
Tolias \etal~\cite{GAJ13}    & 65k & - & 88.0 & 87.9 & \bf{85.0} \\
Tolias \etal~\cite{GJ15}     & 65k & - & -    & \bf{89.4} & 84.0 \\
Razavian~\etal~\cite{RSMC15} & -   & - & \bf{89.7$^*$} & 84.4$^*$ & - \\
\hline
\end{tabular}}%}
\end{table}

\subsection{Visualizing weights}
\label{sec:vis}

\begin{figure}[!h]
\hspace{-4mm}
\begin{tabular}{c}
\vspace{-5mm}
\includegraphics[width=1\linewidth]{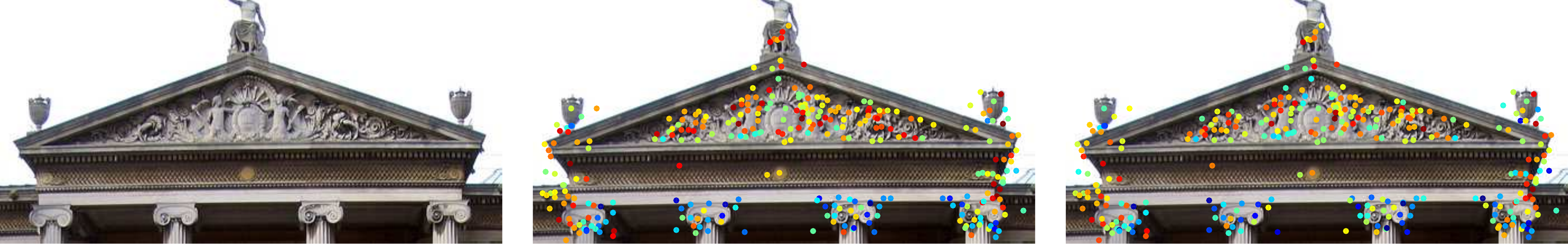} \\
\includegraphics[width=1\linewidth]{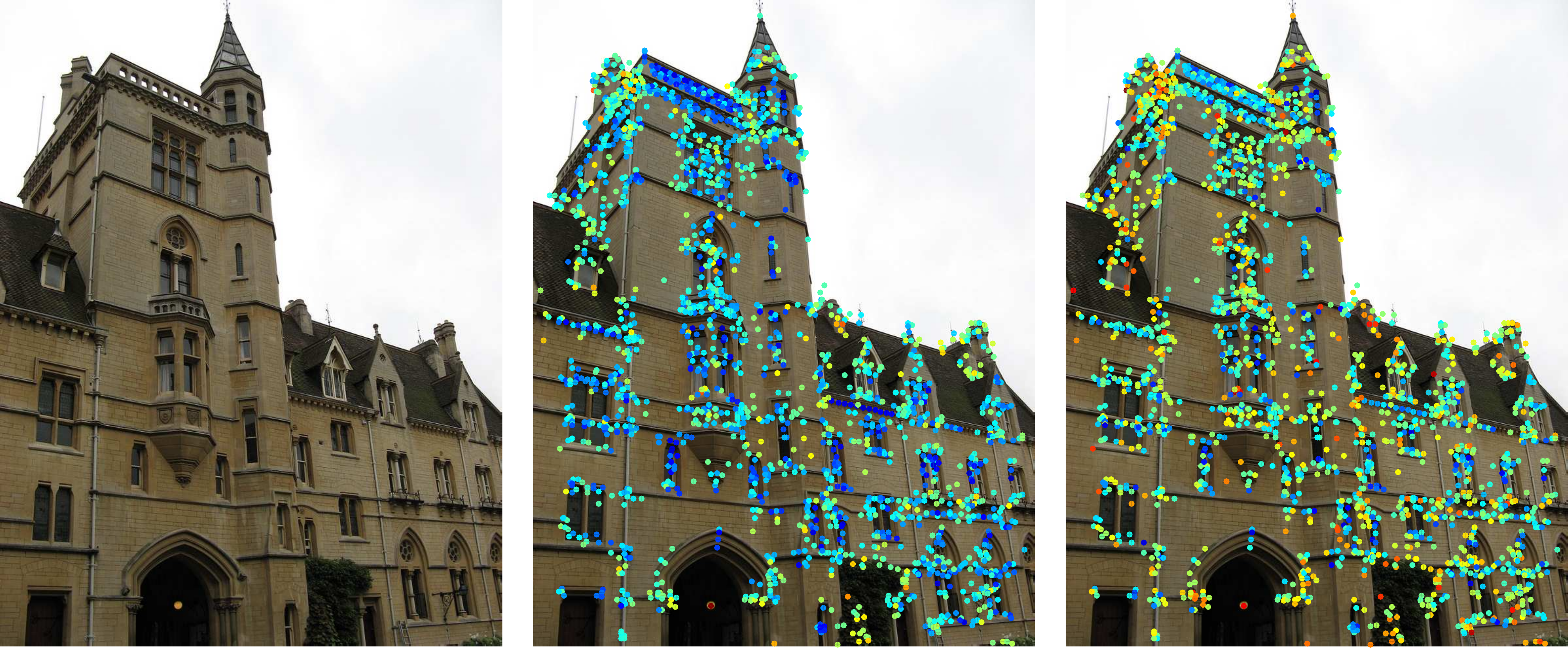} \\
\includegraphics[width=1\linewidth]{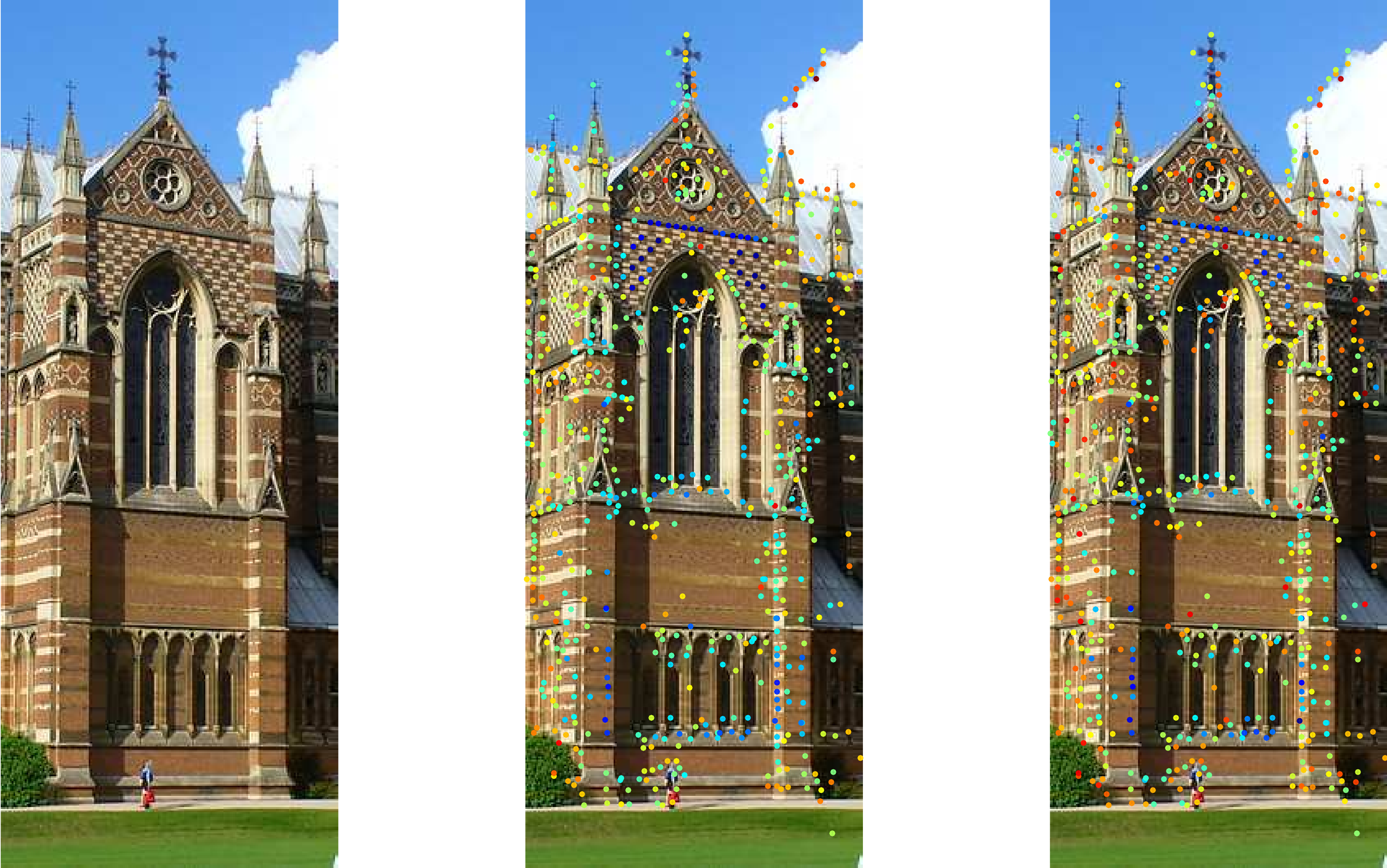} \\
\includegraphics[width=1\linewidth]{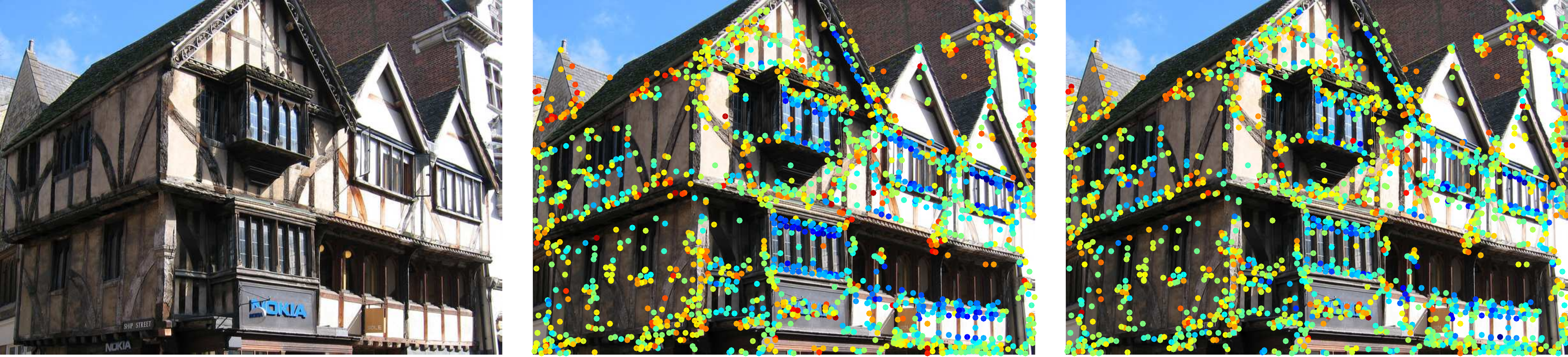} \\
\includegraphics[width=1\linewidth]{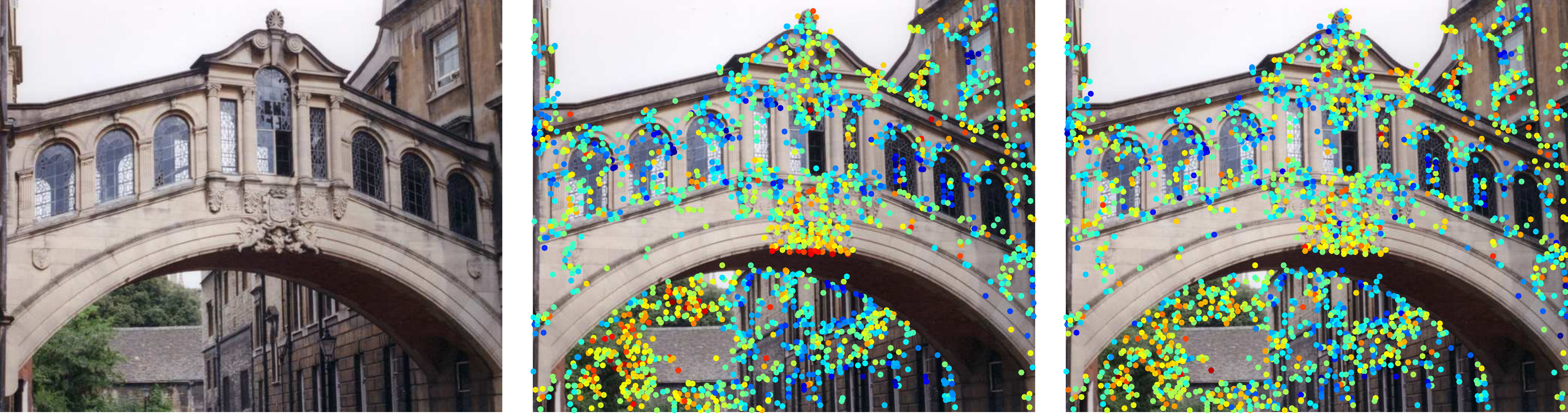} \\
\includegraphics[width=1\linewidth]{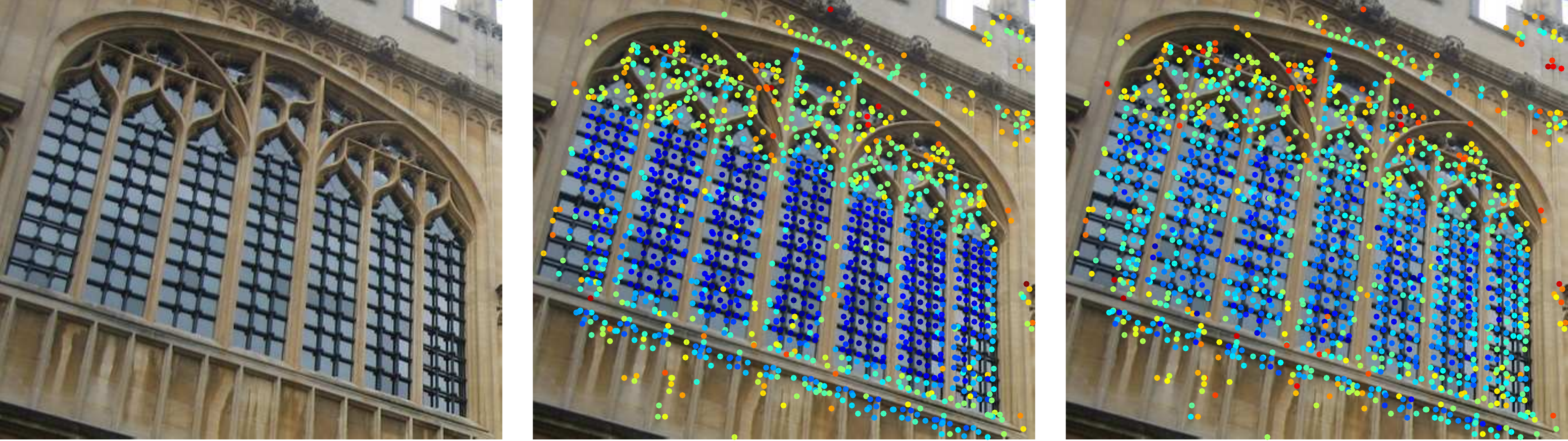} \\
\includegraphics[width=1\linewidth]{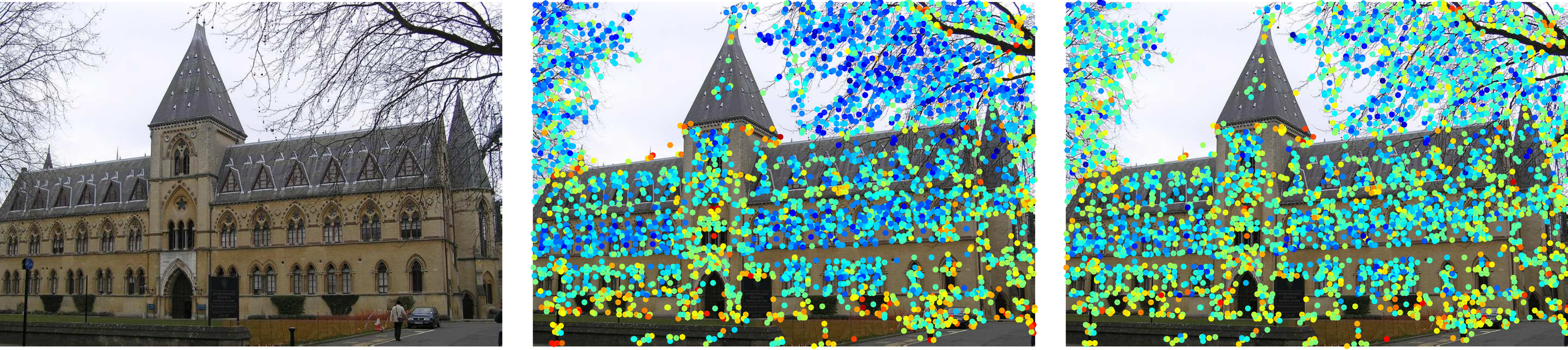} \\
\includegraphics[width=1\linewidth]{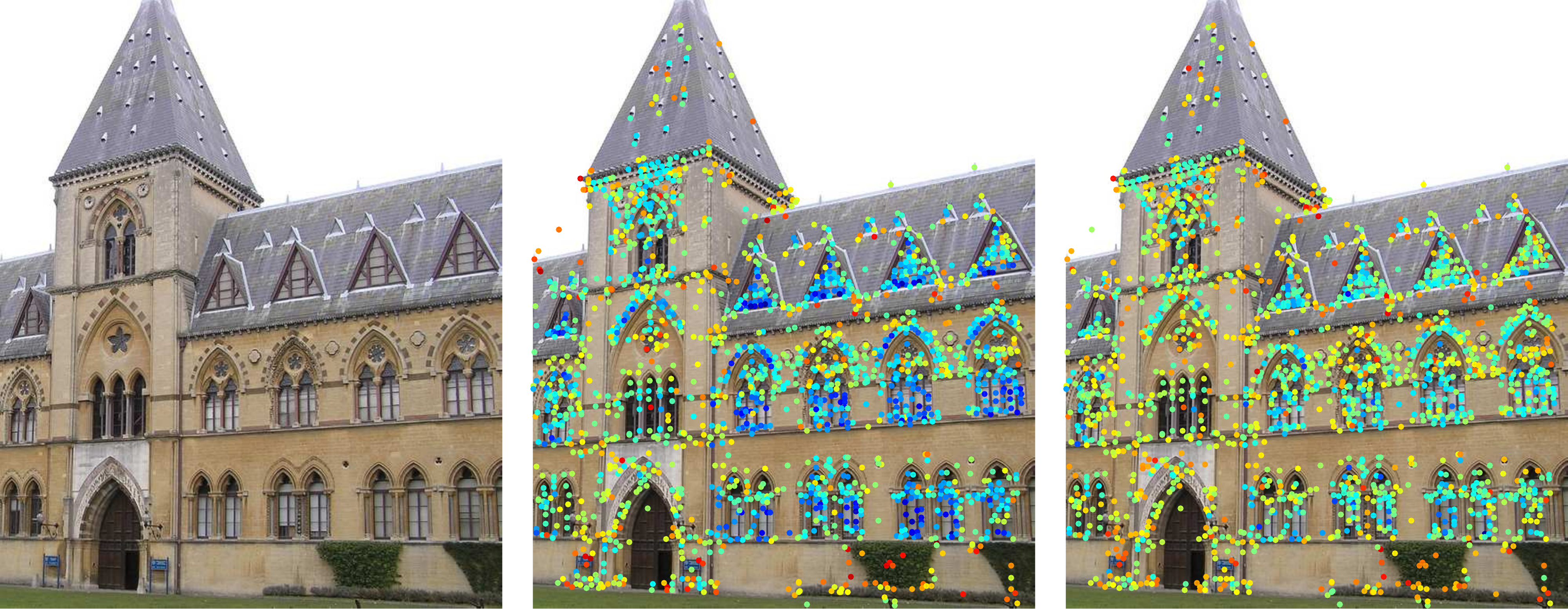} \\
\end{tabular}
\caption{Relative weights for descriptors (warmer colours indicate higher descriptor weights), for a sample of the Oxford query images.
Left: database image. Middle: democratic weights. Right: GMP weights.}
\label{fig:map_egs}
\end{figure}

Figure~\ref{fig:map_egs} illustrates the relative weights of extracted descriptors for a random set of images from the Oxford query image set for $\psid$ and $\psig$ aggregation.
Both methods give low relative weights to descriptors corresponding to repetitive structures such as those found in building facades, but also for foliage (as shown in the penultimate example), which also has repetitive texture.
However, bursty descriptors are more highly penalised for $\psid$ aggregation.

\subsection{Complexity analysis}

\begin{table}[t]
\caption{CPU timings (in seconds) for generating representations.
We do not include the cost of extracting the SIFT descriptors.
\label{tab:timings}}
\centering{
\begin{tabular}{|c|r|rrr|}
\hline
embedding & $\k$ & \multicolumn{3}{c|}{Aggregation} \\
 &  & $\psi_s$ & $\psi_d$ & $\psi_{gmp}$\\
\hline
$\triemb$ & 8	&  0.009 &   0.681 &   0.927 \\
$\triemb$ & 16	&  0.013 &   1.339 &   1.507 \\
$\triemb$ & 32  &  0.026 &   3.351 &   3.758 \\
$\triemb$ & 64  &  0.057 &   8.626 &   9.867 \\
$\triemb$ & 128 &  0.109 &  68.812 &  78.490 \\

\hline
$\fvemb$ & 8  & 0.089 & 0.188 & 0.447 \\
$\fvemb$ & 16 & 0.089 & 0.190 & 0.467 \\
$\fvemb$ & 32 & 0.096 & 0.199 & 0.490 \\
$\fvemb$ & 64 & 0.112 & 0.198 & 0.493 \\
$\fvemb$ & 128& 0.134 & 0.214 & 0.519 \\
\hline
\end{tabular}}
\end{table}

Table~\ref{tab:timings} reports the timings measured to compute our representations for different vocabulary sizes~$\k$.  The measures are obtained on the query images of Oxford5k, and are carried out on an Intel Xeon E5-2680/2.50GHz with 24 cores. We report the CPU times (larger than elapsed ones because CPU time cumulates all active threads). On a quad-core laptop with multi-threading, the timing is typically 20ms per image for $\triemb+\psis$.

Computing $\triemb$ or $\fvemb$ is fast. The bottleneck is weighted aggregation, when adopted, and in particular the computation of the kernel matrix $\K$.
Note that $\fvemb$ used hard assignment resulting in embeddings whose sparsity was exploited to efficiently compute $\K$.
This leads to a significant speed-up in computing the representation, compared to when $\triemb$ is used.
Note also that $\triemb$ was computed with little code optimisation: aggregation is done in plain Matlab, while we have optimised the computation of $\triemb$ with a mex file. This also suggests that further optimisation strategies should be considered for larger vocabularies. A simple effective one is to threshold the gram matrix by setting to 0 all values below a threshold (typically, 0.1), to make it sparse at a small accuracy cost.

Computing $\psid$ is faster than computing $\psig$ due to the fact that the Sinkhorn algorithm is terminated after 10 iterations while the conjugate gradient descent algorithm typically requires around 100 iterations to converge.

